\documentclass[10pt,twocolumn,letterpaper]{article}

\usepackage{cvpr}              

\usepackage{graphicx}
\usepackage{amsmath}
\usepackage{amssymb}
\usepackage{booktabs}
\usepackage{tabularx}
\usepackage{subcaption}
\usepackage{capt-of}
\usepackage{xcolor}
\usepackage{bm}
\usepackage{multirow}
\usepackage{makecell}
\usepackage{siunitx}
\usepackage{gensymb}
\usepackage[accsupp]{axessibility}

\usepackage[pagebackref,breaklinks,colorlinks]{hyperref}

\newcommand{\boldstart}[1]{\noindent\textbf{#1}}
\newcommand{\suppress}[1]{}

\newcommand{\methodname}{MobileBrick\xspace}

\usepackage[capitalize]{cleveref}
\crefname{section}{Sec.}{Secs.}
\Crefname{section}{Section}{Sections}
\Crefname{table}{Table}{Tables}
\crefname{table}{Tab.}{Tabs.}

\def\cvprPaperID{4459} 
\def\confName{CVPR}
\def\confYear{2023}

\begin{document}

\title{\methodname: Building LEGO for 3D Reconstruction on Mobile Devices}

\author{
    Kejie Li\textsuperscript{1},
    Jia-Wang Bian\textsuperscript{1},
    Robert Castle\textsuperscript{2},
    Philip H.S. Torr\textsuperscript{1},
    Victor Adrian Prisacariu\textsuperscript{1} \\
    \textsuperscript{1}University of Oxford,
    \textsuperscript{2}Apple
}

\twocolumn[{%
\renewcommand\twocolumn[1][]{#1}%
\maketitle
\begin{center}
    \centering
    \includegraphics[width=0.9\linewidth]{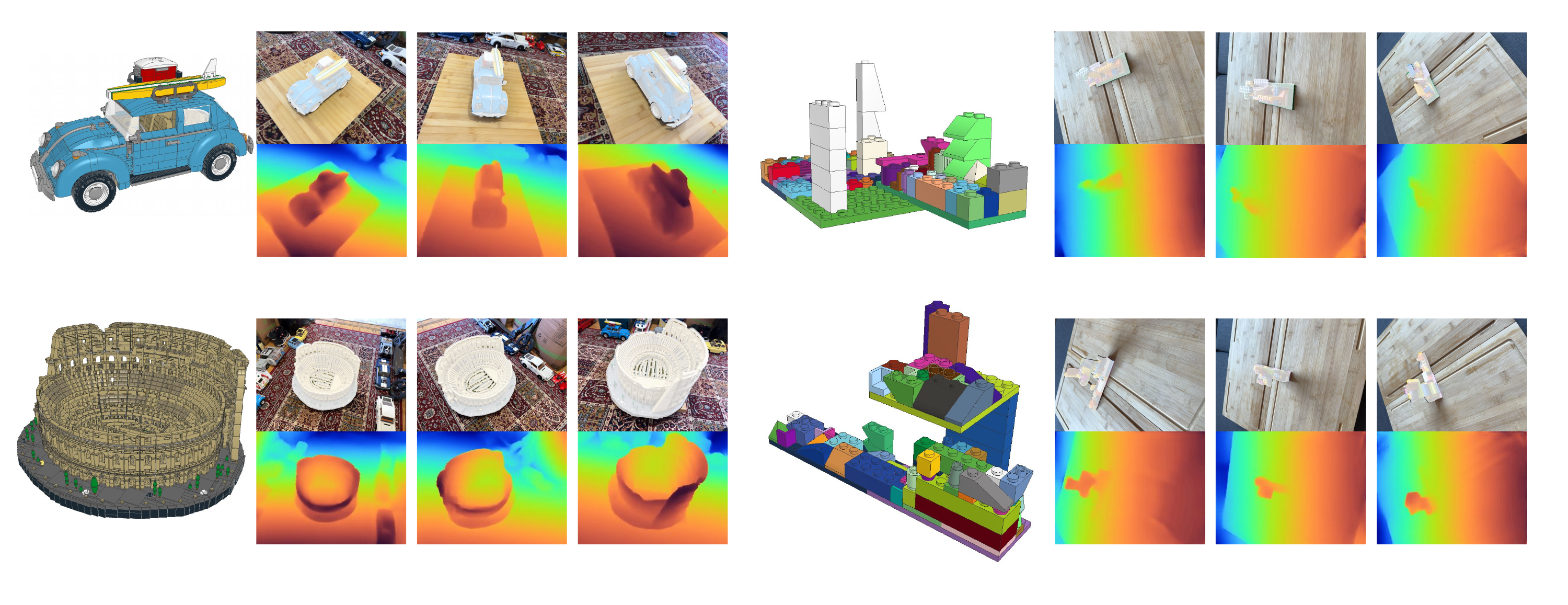}
    \captionof{figure}{Sample LEGO models and RGBD images in the \methodname dataset. The left and right column shows instances in the ``real-world model set'' and the ``random model set'' respectively. We overlay the 3D models after pose alignment to RGB images. }\label{fig:teaser}
\end{center}
}]

\begin{abstract}
High-quality 3D ground-truth shapes are critical for 3D object reconstruction evaluation. 
However, it is difficult to create a replica of an object in reality, and even 3D reconstructions generated by 3D scanners have artefacts that cause biases in evaluation.
To address this issue, we introduce a novel multi-view RGBD dataset captured using a mobile device, which includes highly precise 3D ground-truth annotations for 153 object models featuring a diverse set of 3D structures.
We obtain precise 3D ground-truth shape without relying on high-end 3D scanners by utilizing LEGO models with known geometry as the 3D structures for image capture.
The distinct data modality offered by high-resolution RGB images and low-resolution depth maps captured on a mobile device, when combined with precise 3D geometry annotations, presents a unique opportunity for future research on high-fidelity 3D reconstruction.
Furthermore, we evaluate a range of 3D reconstruction algorithms on the proposed dataset.
\end{abstract}
\vspace{-0.4cm}
\section{Introduction}
High-fidelity 3D object reconstruction from images has always been a ``holy grail'' in computer vision.
It is also the key to enabling immersive and realistic Augment Reality applications.
For instance, a virtual object would not blend in the physical environment realistically even if it is a few millimetres off because the lighting is reflected incorrectly.
Many approaches (\eg, Visual SLAM~\cite{davison2007monoslam}, Depth Fusion~\cite{newcombe2011kinectfusion}, Multi-View Stereo~\cite{schonberger2016structure}) have been proposed to address this problem.
Neural fields have also emerged as a promising technique for 3D reconstruction~\cite{yariv2021volume, wang2021neus} and novel view synthesis~\cite{mildenhall2020nerf}.
Advances in most of these algorithms are made possible with reliable datasets for benchmarking.

\begin{figure}[t]
    \centering
    \includegraphics[width=\linewidth]{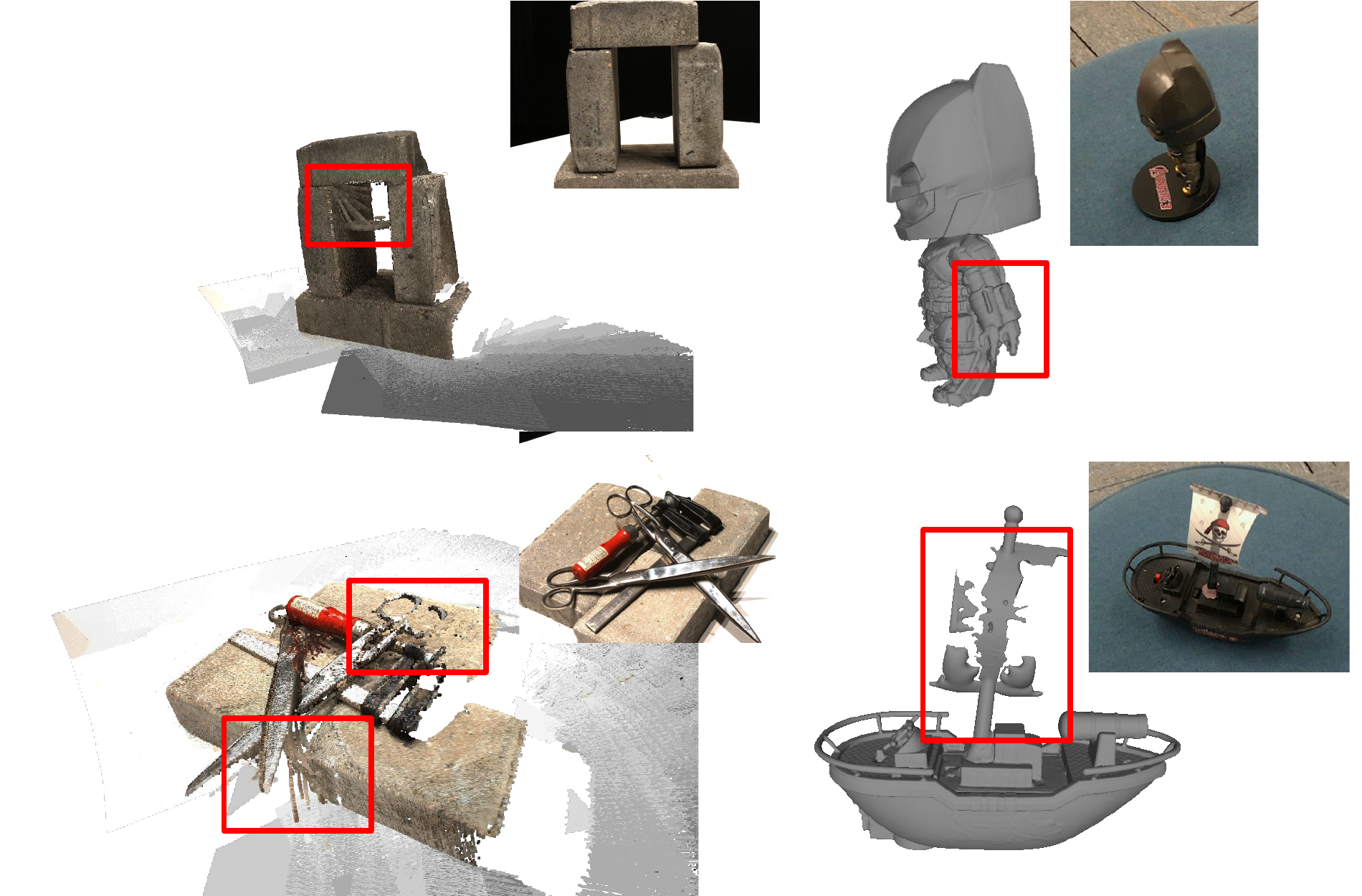}
    \caption{Evaluation references provided by DTU~\cite{jensen2014large} and Shrestha \etal~\cite{shrestha2022real}, shown on the left and right column respectively. Artefacts can be observed despite using a high-end 3D scanner to reconstruct the models. An RGB image for reference is at the top-right corner of each sample.}
    \label{fig:failure}
\vspace{-0.2cm}
\end{figure}

However, building the \emph{exact} 3D model of an object for benchmarking is extremely difficult. 
Some datasets~\cite{jensen2014large,Knapitsch2017,shrestha2022real} resort to high-end 3D scanners to create pseudo-ground-truth models for evaluation, but these 3D models still suffer from artefacts due to noisy measurements, as shown in~\cref{fig:failure}.
Therefore, the DTU dataset~\cite{jensen2014large}, a widely-used dataset for multi-view stereo, is reluctant to call their reconstructed models \emph{ground truth}, but chose the term ``evaluation reference''.
This deviation from the actual ground-truth shape can cause significant biases when evaluating high-fidelity object reconstruction, where millimetres matters.

In addition to the lack of 3D ground-truth, most existing datasets in multi-view reconstruction are not captured on mobile devices, which possess a distinctive data modality from other devices. 
Apple equipping its iPhones and iPads with a compact LiDAR scanner marks a new level of accessibility of depth information alongside RGB information after the rise of commodity-level depth sensors (\eg Kinect and Intel RealSense).
However, the LiDAR scanner on these mobile devices provides low-quality depth maps compared to those of commodity-level sensors.
How to effectively use low-quality yet easily accessible depth maps in 3D object reconstruction is largely unexplored due to the lack of data.


In this paper, we introduce the \methodname dataset, which contains a large collection of object-centric video clips featuring a diverse set of 3D structures.
We use an iOS device with a LiDAR scanner~\cite{apple-lidar} (\eg iPhone 13 Pro, iPad Pro) as an image capture device that can provide calibrated high-resolution RGB images accompanying low-resolution depth maps.
The camera poses of images are tracked using ARKit~\cite{apple-arkit}.

More importantly, we provide the exact 3D model of the structure that is aligned to the image sequence without relying on high-end scanning devices.
The key to acquiring the ground-truth 3D model for the object of interest in a video is LEGO models.
The fact that LEGO models are modular structures connected by primitive and well-defined bricks allows us to acquire the exact 3D ground-truth shape\footnote{subject to manufacturing tolerance of LEGO bricks, which is as small as 0.01 millimetres} with no accuracy loss in measurement. 

In order to align the 3D ground-truth shape to the image sequence, we design a human-in-the-loop pose alignment procedure with three steps. First, we align the 3D model to a single image in the sequence by solving a Perspective-n-Point problem with annotated keypoints.
Second, we verify and refine the alignment by maximising the overlap between the 3D model's projections on a few sampled images and its observed location in those images. 
Third, we use bundle adjustment to refine the camera poses of all images in the sequence such that they are consistent with the aligned object pose and to alleviate motion drift caused by ARKit.


Overall, our contributions in this paper are as follows:
\begin{itemize}
    \item We propose \methodname, a large-scale dataset of $153$ diverse object shapes focusing on detailed 3D object reconstruction with a unique data modality of high-resolution RGB images with low-resolution depth maps captured on a mobile device. 
    \item We provide exact ground-truth 3D models by building the digital replica of each object and we design an efficient annotation pipeline to align the 3D models to image sequences.
    \item We demonstrate the usefulness of the proposed dataset by training and evaluating various methods on the tasks of multi-view surface reconstruction, novel view synthesis, and colour-guided depth enhancement.
\end{itemize}


\begin{table*}[t]
    \centering
    \small
    \begin{tabular}{c | c c | c c c c c c}
    & Redmond-OS~\cite{Choi2016} & CO3D~\cite{reizenstein2021common} & DTU~\cite{jensen2014large} & T \& T~\cite{knapitsch2017tanks} & Shrestha \etal~\cite{shrestha2022real} & Pix3D~\cite{sun2018pix3d} & Ours \\
    \hline
    \hline
    Exact 3D model
    & $\times$ & $\times$ & $\star$ & $\star$ & $\star$ & $\checkmark$ & $\checkmark$ \\
    Mobile depth maps 
    & $\times$ & $\times$ & $\times$ & $\times$ & $\times$ & $\times$ & $\checkmark$ \\
    Multiple views 
    & $\checkmark$ & $\checkmark$ & $\checkmark$ & $\checkmark$ & $\checkmark$ & $\times$ & $\checkmark$ \\
    Models/Scenes 
    & 100 & 19k & 80 & 14 & 900 & 300 & 153
    \end{tabular}
    \caption{Comparison between different datasets. Redmond-OS and CO3D provide 3D models that are reconstructed by TSDF-Fusion and COLMAP respectively, which are too inaccurate to be used in 3D reconstruction benchmark. DTU, Tanks and Temples, and Shrestha \etal~\cite{shrestha2022real} resort to 3D scanners to capture higher-quality 3D models, but still not error-free (hence denoted as $\star$). Only Pix3D and ours provide exact 3D models, but Pix3D has only a single image associated to each 3D model. Furthermore, we are the only dataset that provides depth maps captured on a mobile device.}
    \label{tab:dataset_comparison}
    \vspace{-0.2cm}
\end{table*}

\section{Related Work}
\subsection{Datasets for Multi-view Reconstruction}
DTU~\cite{jensen2014large} and Tanks and Temples~\cite{knapitsch2017tanks} datasets are often used to benchmark multi-view reconstruction algorithms in the computer vision community. 
While they acquire 3D models using high-end 3D scanners,
the models are subject to artefacts, as shown in~\cref{fig:failure}. 
In contrast, we align the exact 3D ground-truth model to the image sequence in \methodname.
Furthermore, devices used for data collection in these datasets are notably different from the mobile devices we use nowadays.
We provide RGBD images whereas they have RGB images only.

Some datasets~\cite{reizenstein2021common,ahmadyan2021objectron} comprise a significantly larger number of sequences or models, which are particularly useful for training deep neural networks.
However, the lack of accurate and dense 3D ground-truth 
 hampers their use to benchmark reconstruction algorithms.
While synthetic datasets~\cite{chang2015shapenet,mccormac2016scenenet,roberts2021hypersim,fu20213d} offer a large number of images associated with exact 3D models, the domain gap between real and synthetic images causes a significant discrepancy when testing on real images~\cite{zhao2019geometry}.

Aligning a 3D CAD model to images, as a way of generating 3D ground truth, has been explored in previous object-centric datasets. 
Pascal3D~\cite{xiang2014beyond} and ObjectNet3D~\cite{xiang2016objectnet3d} ask annotators to pair images with an object in a pre-defined object database.
This annotation process cannot guarantee that an object that appears in an image is matched exactly unless the object is also included in the object database.

To solve this issue, Pix3D~\cite{sun2018pix3d} collects images of IKEA furniture exclusively. 
They then align an IKEA CAD model to each image to acquire the precise 3D ground truth.
Because they align a CAD model to a single image, they are limited to evaluate single-view reconstruction mainly driven by deep learning algorithms, and it is impossible to be used for multi-view reconstruction. 

The closest to our dataset is Shrestha \etal~\cite{shrestha2022real}. Both datasets provide object-centric RGBD sequences with a few key differences as follows.
First, Shrestha \etal~\cite{shrestha2022real} use a 3D scanner to acquire evaluation reference, which shares limitations presented in the DTU~\cite{jensen2014large} and Tanks and Temples~\cite{knapitsch2017tanks} datasets.
Second, while they also provide depth maps alongside the RGB images, they capture high-resolution depth maps using a LiDAR depth camera instead of the one on a mobile device. 
This conceals the challenges that low-resolution depth maps captured by mobile devices bring in 3D reconstruction.

\subsection{Datasets with Depth Maps}
Various RGBD datasets~\cite{dai2017scannet,chang2017matterport3d,Choi2016,silberman2012indoor} for 3D reconstruction and 3D scene understanding have been proposed thanks to the rise of commodity-level depth sensors (\eg Kinect, and Intel RealSense) in the last decade.
However, as noted in ARKitScene~\cite{baruch2021arkitscenes}, the gap in depth sensing technology between the commodity-level depth sensors and mobile devices is so significant that algorithms that are developed using these datasets are difficult to be used on data captured on mobile devices.
Furthermore, most RGBD datasets offer data and annotations for scene-level understanding, while the proposed dataset provides RGBD images captured on a mobile device, with the focus on 3D object reconstruction.

\begin{figure*}[t]
    \centering
    \includegraphics[width=\linewidth]{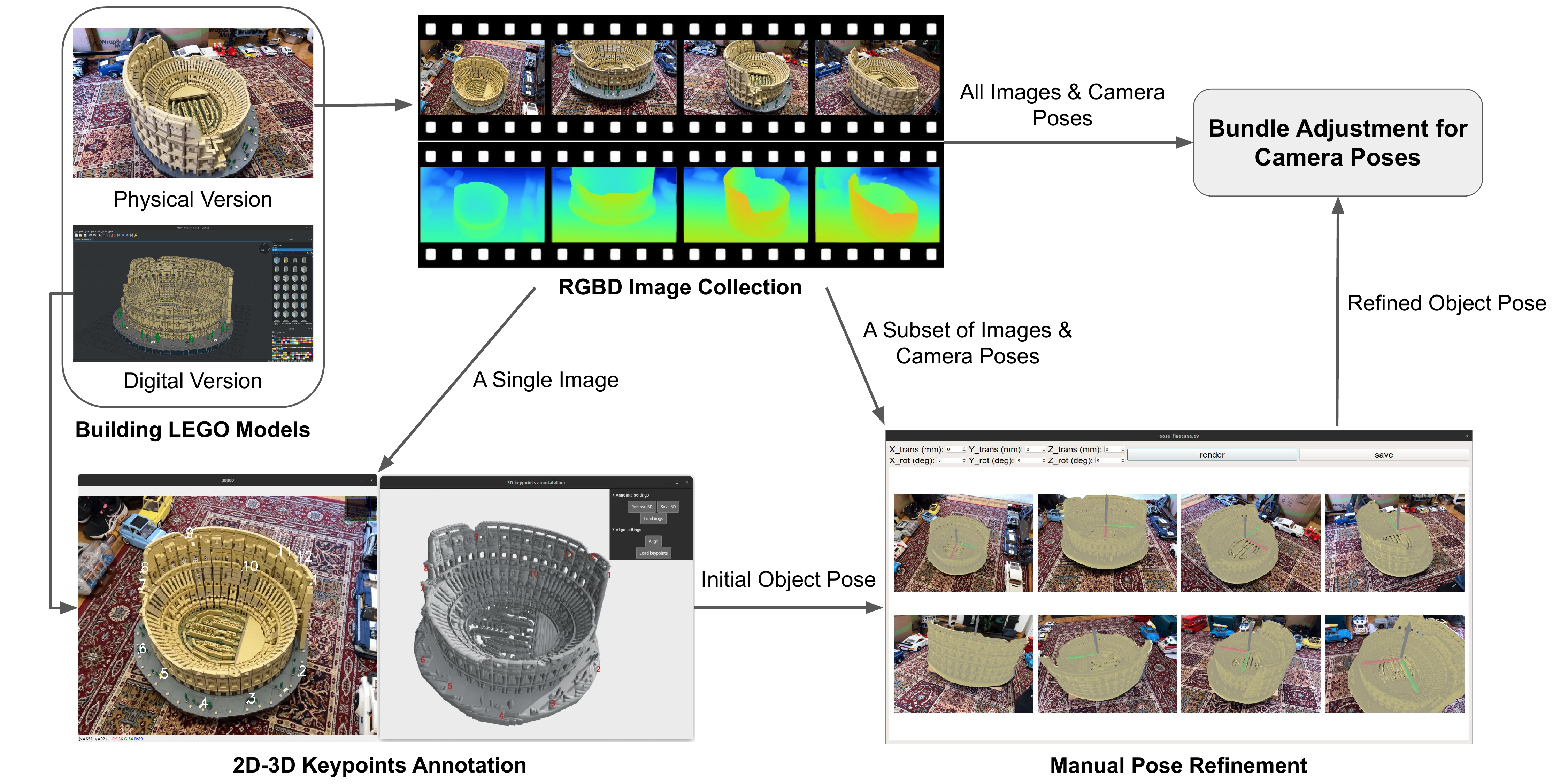}
    \caption{The data collection and annotation pipeline. After image capture, the digital model is first aligned to the world coordinate system by solving the PnP problem using annotated 2D-3D keypoints correspondences. The object pose is further refined manually by maximising the projections of the digital model on the object foreground on a subset of images. Lastly, we finetune all camera poses by bundle adjustment.}
    \label{fig:pipeline}
\end{figure*}
\section{Data Acquisition and Annotation}
We first describe how we capture images given a collection of physical LEGO models and a mobile device (\cref{sec:image_collection}) and how we acquire the digital replica of each model (\cref{sec:3d_model}). 
We then introduce a model alignment procedure, where the 3D models are registered to the sequence coordinate system (\cref{sec:alignment}).
The aligned 3D models are pre-processed for evaluation in ~\cref{sec:preprocess}. 
Finally, we conclude this section with statistics of the dataset (\cref{sec:statistic}). 

\subsection{Image Collection}\label{sec:image_collection}
We record video clips on an iPhone moving around an object of interest.
We maintain a distance of
20-50 cm and move in a circular motion because 1) the low-resolution LiDAR sensor needs to be in proximity to the object to capture geometry details, and 2) the LiDAR sensor fails when it is very close the object.
At around 10 frames per second (fps), ARKit provides high-resolution RGB images at $1440\times1920$ pixels and low-resolution depth maps at $192\times256$ pixels. 
It also outputs both the extrinsic and intrinsic parameters of each frame.

\subsection{3D Model Acquisition}\label{sec:3d_model}
We need both digital and physical 3D models for evaluation and image capturing, respectively.
The modular structure of LEGO models guarantees that the physical and digital version would match exactly if they are built following the instructions strictly.
This is the key to aligning RGBD images of a model to the exact 3D ground-truth shape.
There are two types of 3D models in the dataset that can be categorised as ``real-world model set'' and ``random model set'' depending on how a model is built.

The real-world model set refers to official LEGO models that are created by human designers. 
They are resized replicas of renowned objects or landmarks (\eg, the UK parliament, the Hubble Space Telescope) or fiction (\eg, the Hogwarts Castle in Harry Potter, the Millennium Falcon in Star Wars). 
The digital models in this category are available on LEGO modelling websites\footnote{https://www.mecabricks.com/ and http://omr.ldraw.org/} or are created in BrickLink studio~\cite{bricklink} by following the building instructions.


Loose parts of a model that are hard to align, such as the hands of the clock on Big Ben, are removed.
If misalignments are found, we either modify the digital version to match the physical one or discard this model if such a modification is not feasible. 
\cref{fig:teaser} illustrates several examples of LEGO models in this category.

The number of models in the real-world model set is limited and subject to the accessibility of the physical LEGO models for image collection. 
Therefore, we propose a random model set that comprises a large collection of models that are randomly generated.
Following ~\cite{wang2022translating}, we use procedural generation to create a large amount of digital LEGO models using $72$ primitive LEGO bricks randomly.
After the digital models are created, we use BrickLink Studio~\cite{bricklink} to generate the building instructions for the physical models. 
To build the physical LEGO model, each volunteer is assigned a collection of LEGO bricks that are needed to build the model accompanied by the building instructions. 

A concern of using randomly generated models is whether the knowledge learnt from these models is generalisable to real-world objects.
To this end, we conduct experiments on multi-view reconstruction and depth enhancement to demonstrate that training on the random model set can improve the network performance when testing on the real-world model set.

\subsection{3D Model Alignment}\label{sec:alignment}
The most important procedure for building the proposed dataset is aligning a digital 3D model to an image sequence, so that the aligned model can be used for evaluation.
We propose a three-stage alignment pipeline to achieve highly accurate alignment. 
In the first stage, we formulate the alignment as a Perspective-n-Point (PnP) problem that aligns the object pose to an image in the video using manually annotated 2D-3D keypoint correspondences.
We recognise that it is impossible to achieve pixel-perfect 2D-3D correspondences with human annotations, so this alignment is only a rough estimation.
In the second stage, we manually refine the object pose by minimising the reprojection error from a few sampled viewpoints in the video using a Graphical User Interface (GUI). 
Lastly, to ensure the pose alignment is consistent in all images and alleviate camera drift caused by ARKit, we perform a bundle adjustment to refine all camera poses.

\boldstart{PnP alignment} 
The annotators are asked to annotate $10$-$20$ 3D keypoints and their projections on an image given a user interface---See ``2D-3D Keypoints Annotation'' in~\cref{fig:pipeline}.
They are encouraged to annotate distinctive positions (\eg, corners and studs on LEGO bricks) and avoid only annotating coplanar points, which would cause the PnP algorithm to fail.
In the case where the projections of some 3D keypoints are not visible on the image, they have the option to select another image where all 3D keypoints are visible.

\boldstart{Multi-view manual refinement}
Since acquiring pixel-perfect correspondences is difficult, it is necessary to further refine the object pose.
We provide a user interface that shows the projections of the 3D model using the pose from the first stage on $8$ different images sampled uniformly in the image sequence (visualised in ``Manual Pose Refinement'' in~\cref{fig:pipeline}).
Note that we do not need to repeat the PnP step for these $8$ images because relative camera poses are tracked by ARKit.
The annotators can adjust the $6$ Degree-of-Freedom (DoF) transformation parameterised by $3$ DoF translation and $3$ DoF Euler angles, such that the overlap between the model projections and the object foreground in the images is maximised.

\boldstart{Bundle adjustment}
After the aforementioned steps, the object is aligned well with a subset of image frames that are used for manual refinement. 
To propagate the object alignment to the rest of the sequence and rectify the camera drift caused by ARKit, we refine the camera poses by minimising reprojection errors.
We first extract features from all images and match them across neighbouring images. 
Unlike a normal structure from motion problem where the 3D geometry is also unknown, we take advantage of the aligned 3D model to 1) determine the 3D position of each feature point (instead of triangulation), 2) determine its visibility from other images, and 3) filter out incorrect matchings.
Specifically, if the Euclidean distance between the 3D points unprojected from two matched pixels is larger than $2cm$, this matching is discarded.
After removing outliers, the camera poses $\mathbf{P}_1, \mathbf{P}_2, ..., \mathbf{P}_n$ of $n$ images are refined by minimising the following cost function:
\begin{equation}
    F(\mathbf{P}_1, \mathbf{P}_2, ..., \mathbf{P}_n) = \sum_{i=1}^n \sum_{j=1}^m \| \mathbf{x}_{ij} - Proj(\mathbf{X}_j, \mathbf{P}_i, \mathbf{K}_i)\|^2,
\end{equation}
where $\mathbf{X}_j$ denotes the $j^{th}$ 3D keypoint, $\mathbf{x}_{ij}$ is the observed position of the $j^{th}$ keypoint from the $i^{th}$ image, $\mathbf{K}_i$ is the intrinsic matrix of the $i^{th}$ camera, and $Proj(\cdot)$ describes the rigid transformation and perspective projection.

\begin{figure}
    \centering
    \includegraphics[width=\linewidth]{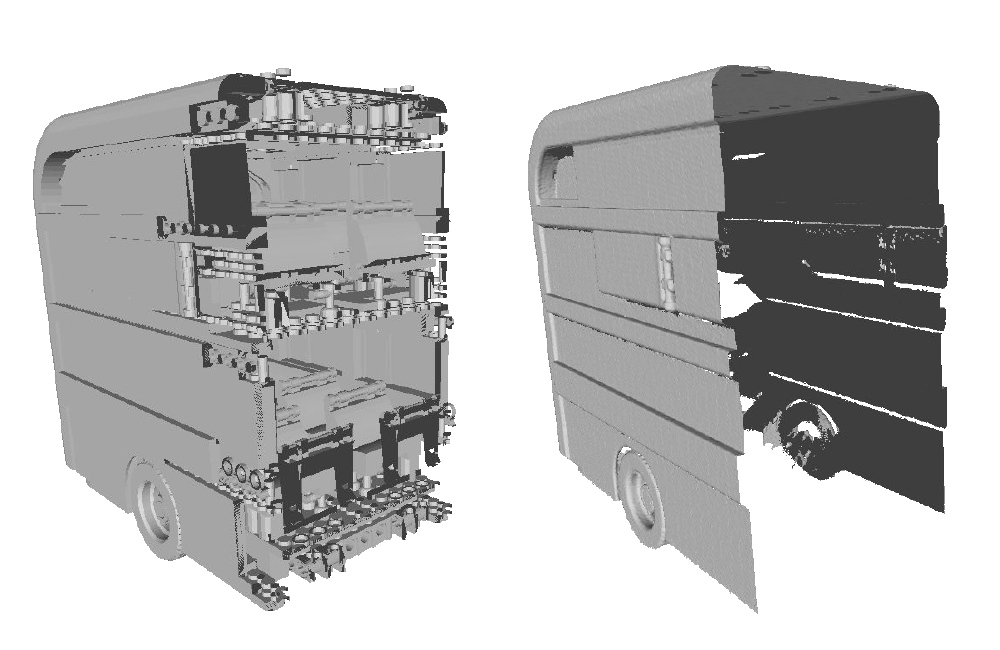}
    \caption{The left shows the native mesh provided by the LEGO modelling website. The right is the processed mesh for evaluation. Note that surfaces that are invisible from images are removed.}
    \label{fig:preprocess}
    \vspace{-0.4cm}
\end{figure}

\begin{table*}[t]
    \centering
    \small
    \begin{tabular}{c c | c c c | c c c | c}
    \multirow{2}{*}{Methods} & \multirow{2}{*}{Input} & & $\sigma=2.5mm$ & & & $\sigma=5mm$ & & \multirow{2}{*}{Chamfer (mm) $\downarrow$} \\
     & & Accu. (\%) $\uparrow$ & Rec. (\%) $\uparrow$ & F1 $\uparrow$ & Accu. (\%) $\uparrow$ & Rec. (\%) $\uparrow$ & F1 $\uparrow$ &  \\
    \hline
    \hline
    TSDF-Fusion~\cite{zhou2018open3d} & Depth & 42.07 & 22.21 & 28.77 & 73.46 & 42.75 & 53.39 & 13.78\\
    BNV-Fusion~\cite{li2022bnv} & Depth & 41.77 & 25.96 & 33.27 & 71.20 & 47.09 & 55.11 & 9.60 \\
    Neural-RGBD~\cite{azinovic2022neural} & RGBD & 20.61 & 10.66 & 13.67 & 39.62 & 22.06 & 27.66 & 22.78 \\
    COLMAP~\cite{schoenberger2016sfm} & RGB & 74.89 & 68.20 & 71.08 & 93.79 & 84.53 & 88.71 & 5.26 \\
    Vis-MVSNet~\cite{zhang2020visibility} & RGB & \textbf{79.83} & 47.25 & 58.32 & \textbf{97.35} & 65.90 & 77.49 & 9.27 \\
    \makecell{Vis-MVSNet~\cite{zhang2020visibility} \\ (finetuned)} & RGB & 75.64 & 53.64 & 62.01 & 96.03 & 72.42 & 81.89 & 9.52 \\
    NeRF~\cite{mildenhall2020nerf} & RGB & 47.11 & 40.86 & 43.55 & 78.07 & 69.93 & 73.45 & 7.98 \\
    NeuS~\cite{wang2021neus} & RGB & 77.35 & \textbf{70.85} & \textbf{73.74} & 93.33 & \textbf{86.11} & \textbf{89.30} & \textbf{4.74} \\
    \hline
    \hline  
    \end{tabular}
    \caption{Multi-view reconstruction experiment. NeuS achieves the state-of-the-art performance. Vis-MVSNet (finetuned) outperforms the baseline by a large margin. Compared to neural-field-based approaches, methods using Multi-View Stereo (\eg, COLMAP and Vis-MVSNet) can reconstruct accurately, but suffer from surface coverage as shown in ~\cref{fig:mvs}. }
    \label{tab:mvs_exp}
\end{table*}

\renewcommand\tabularxcolumn[1]{m{#1}}
\begin{figure}[th]
    \centering
    \newcolumntype{Y}{>{\centering\arraybackslash}X}
    \begin{tabularx}{\linewidth}{@{}X@{}Y@{\,}Y@{\,}Y@{}}
        RGB GT &
        \includegraphics[width=\linewidth]{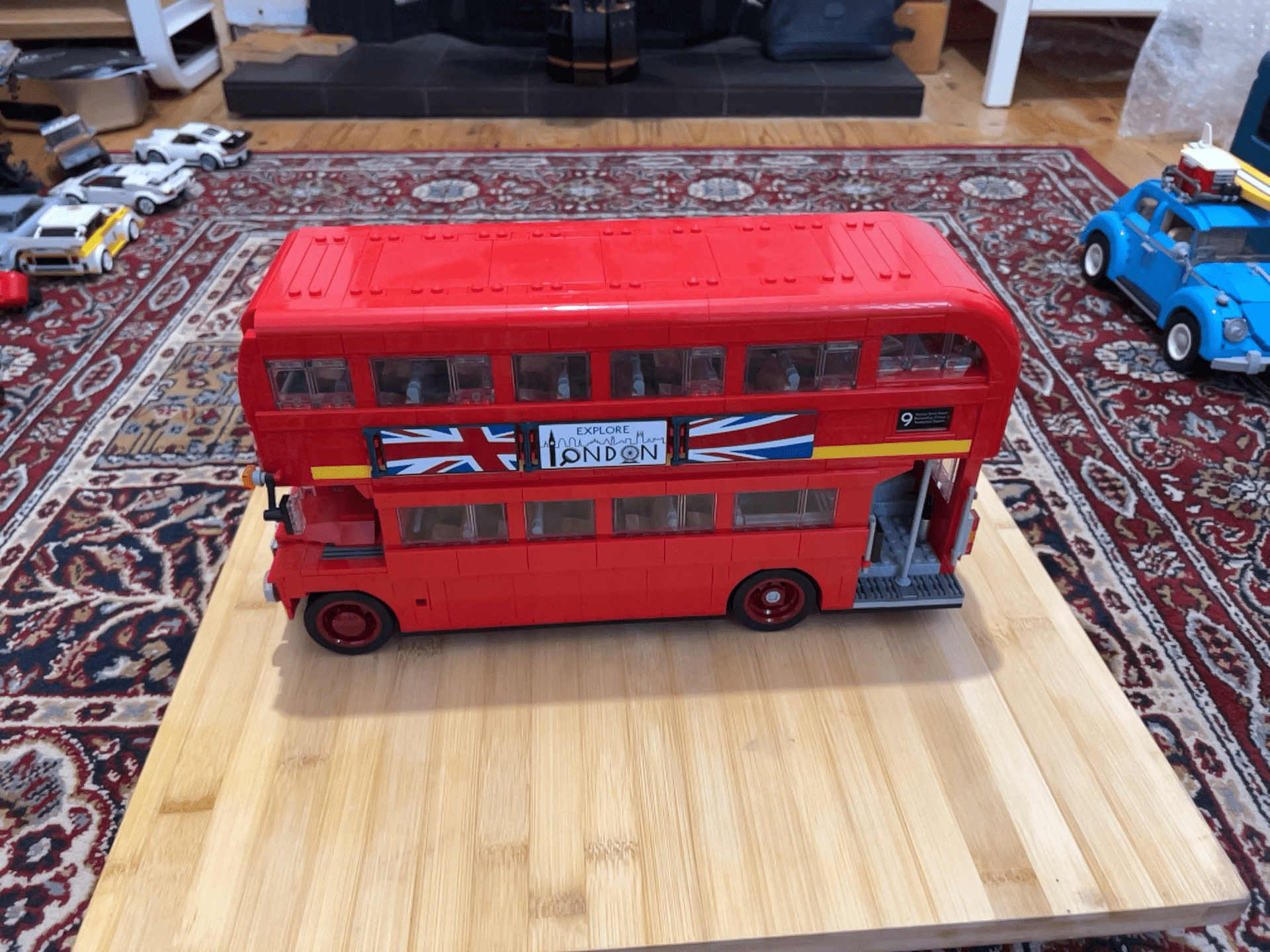}&
        \includegraphics[width=\linewidth]{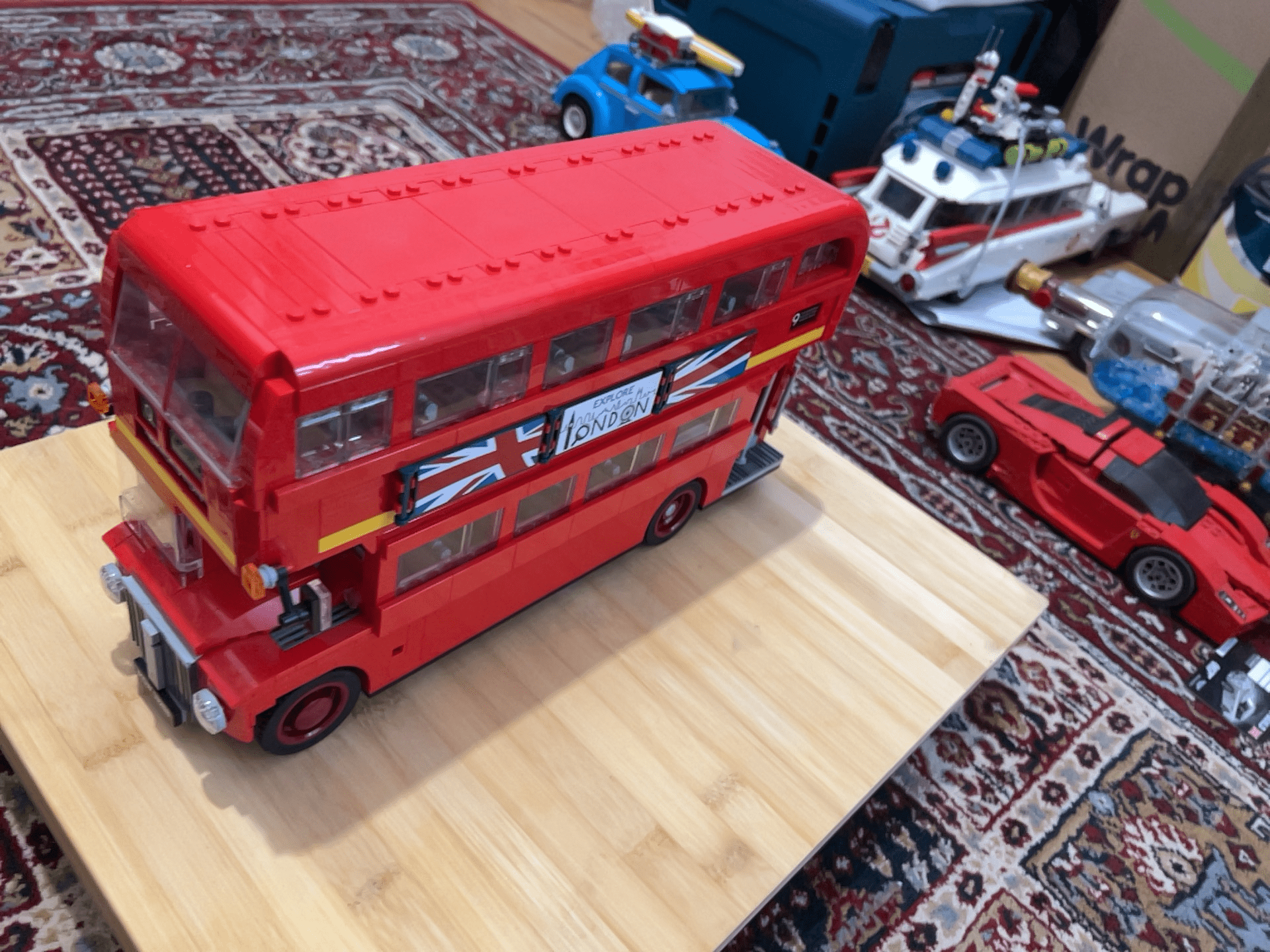}&
        \includegraphics[width=\linewidth]{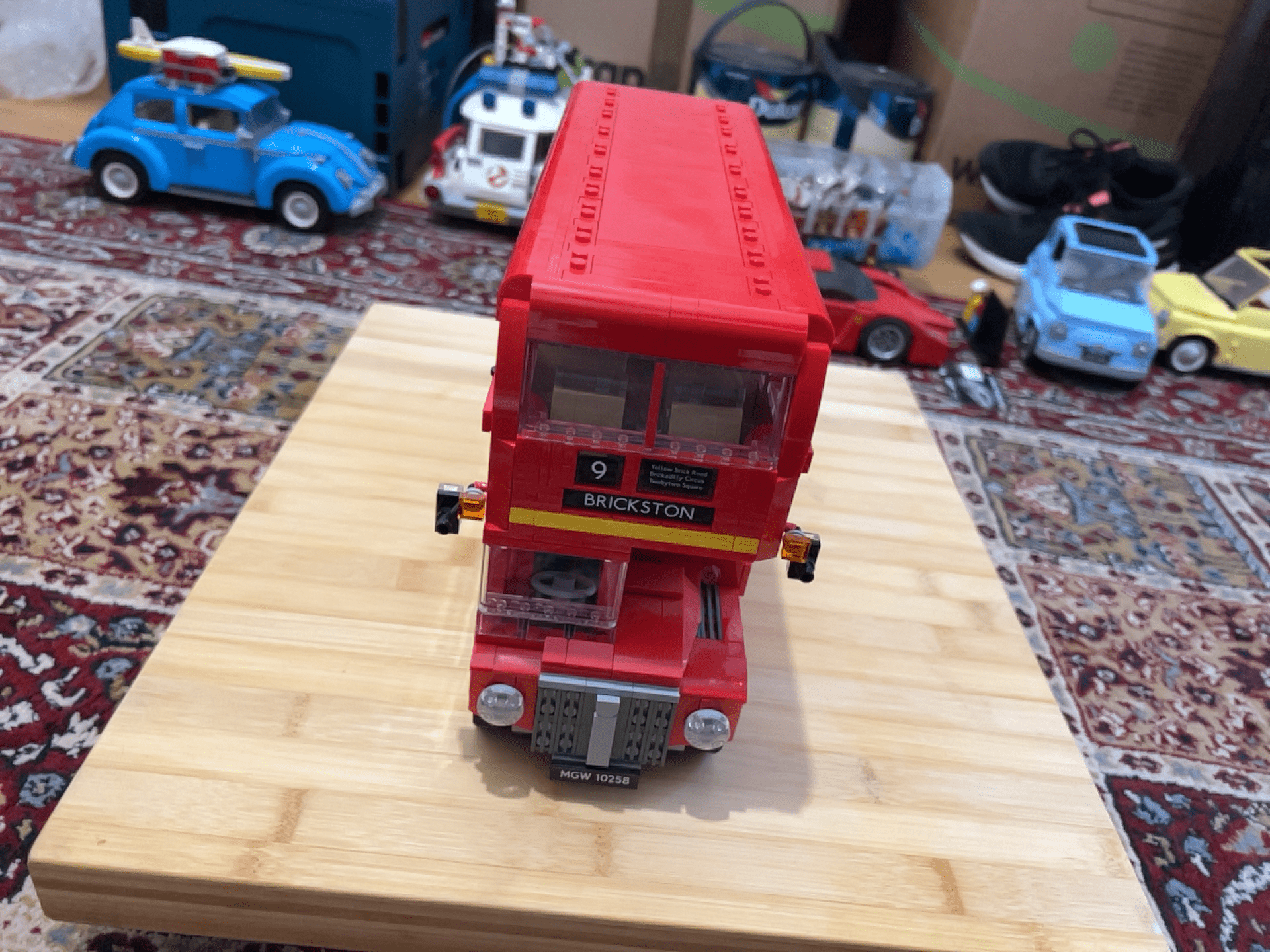} 
        \\
        DVGO &
        \includegraphics[width=\linewidth]{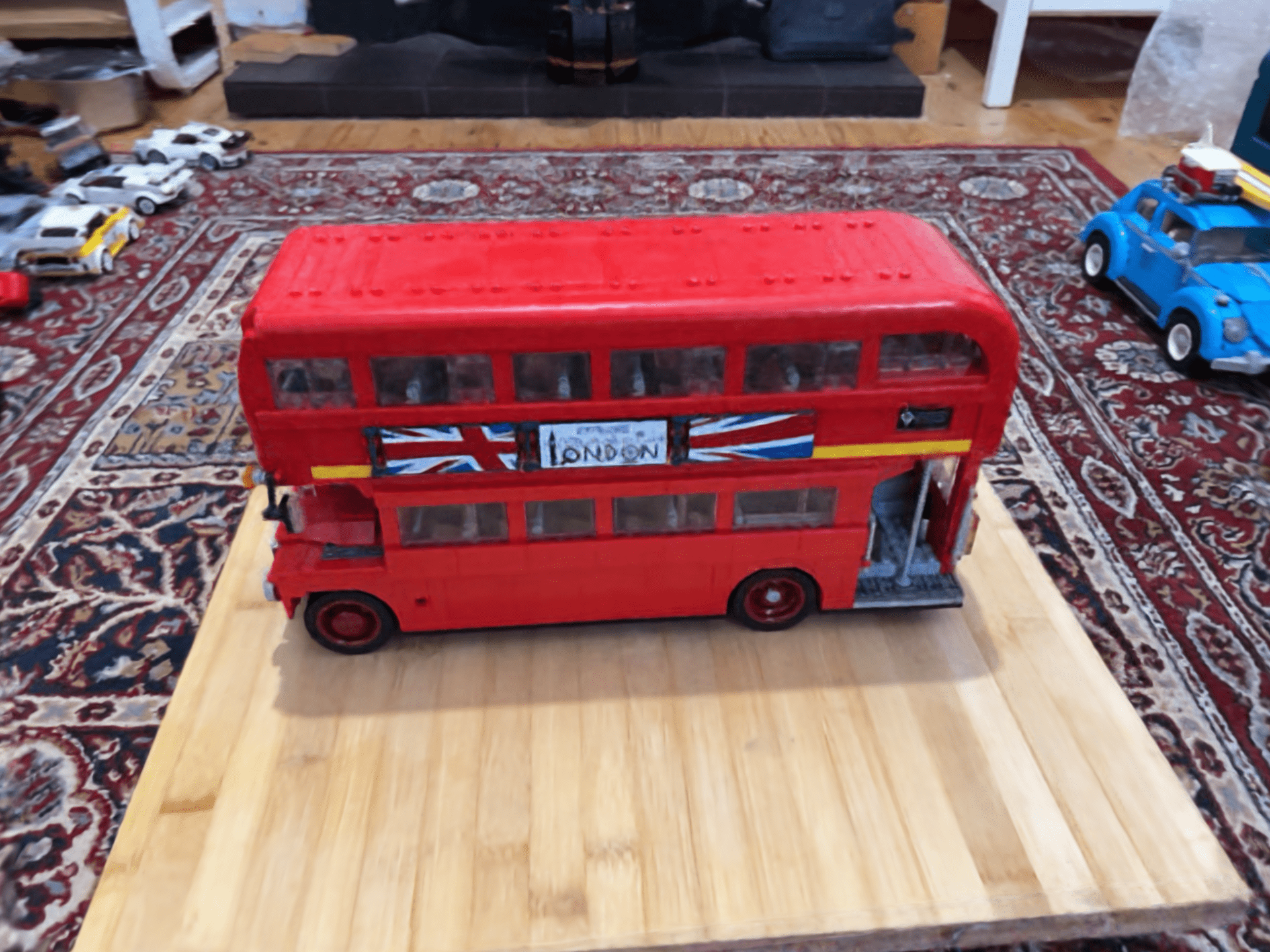}&
        \includegraphics[width=\linewidth]{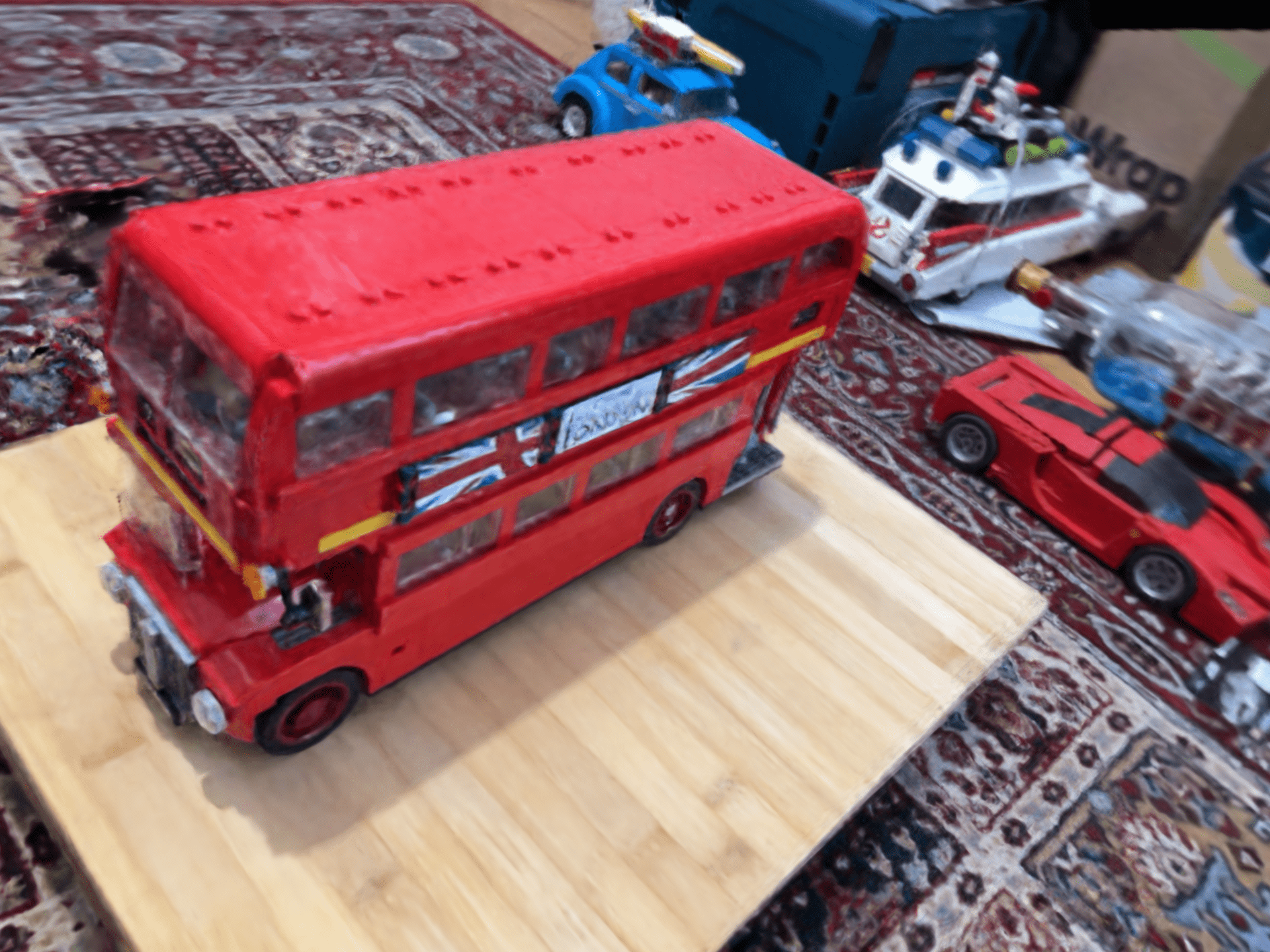}&
        \includegraphics[width=\linewidth]{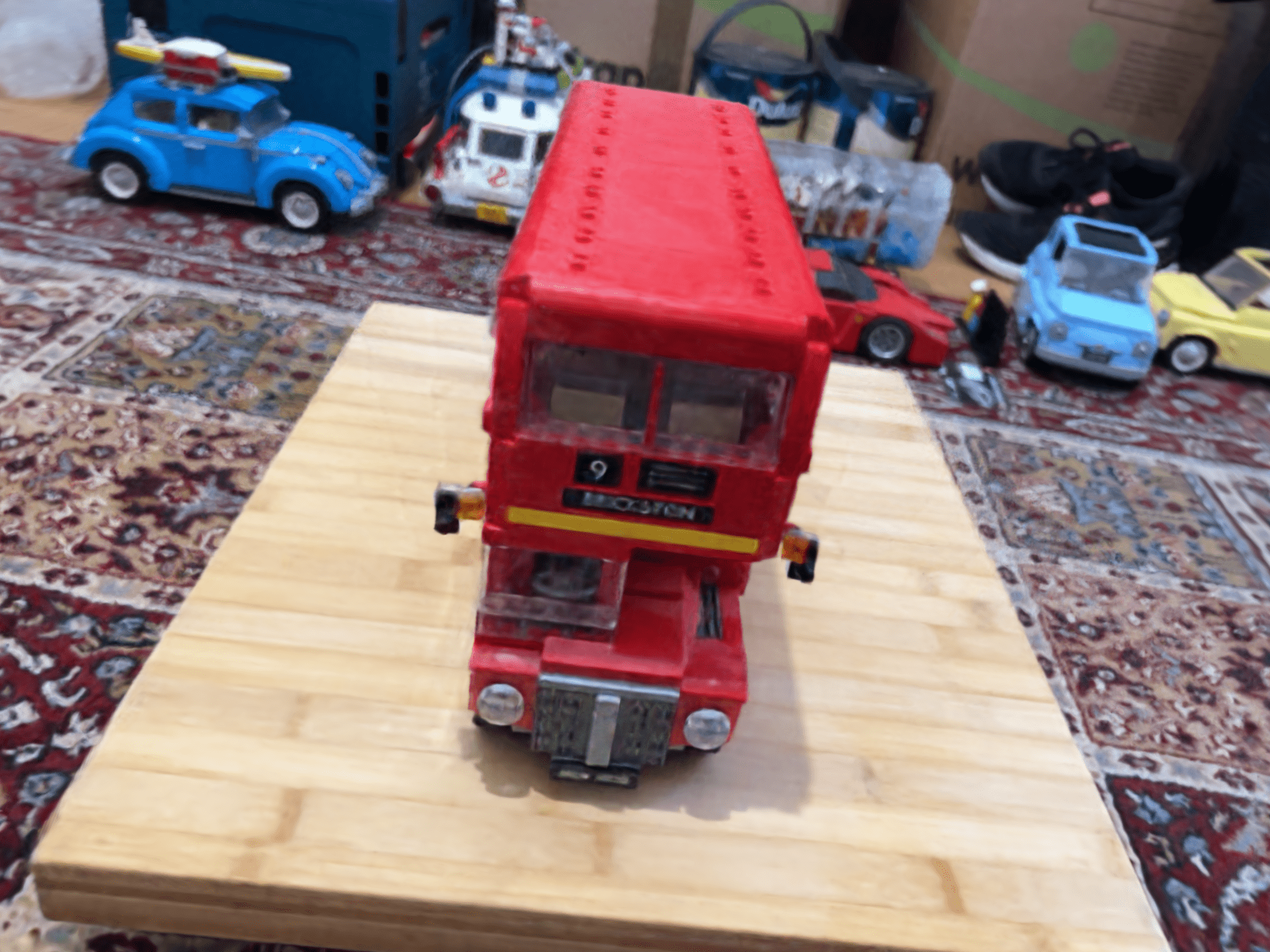} 
        \\
        NeRF &
        \includegraphics[width=\linewidth]{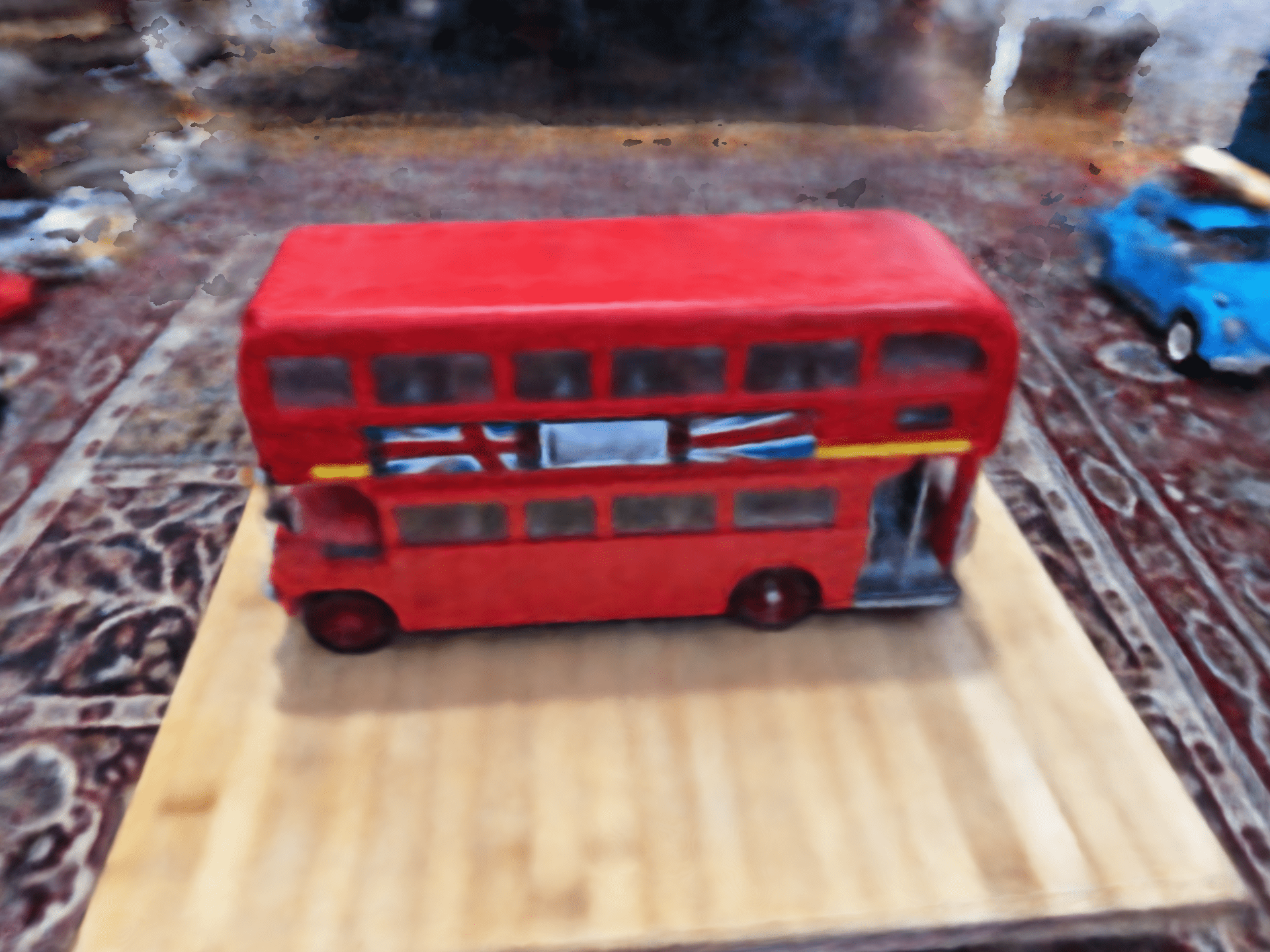}&
        \includegraphics[width=\linewidth]{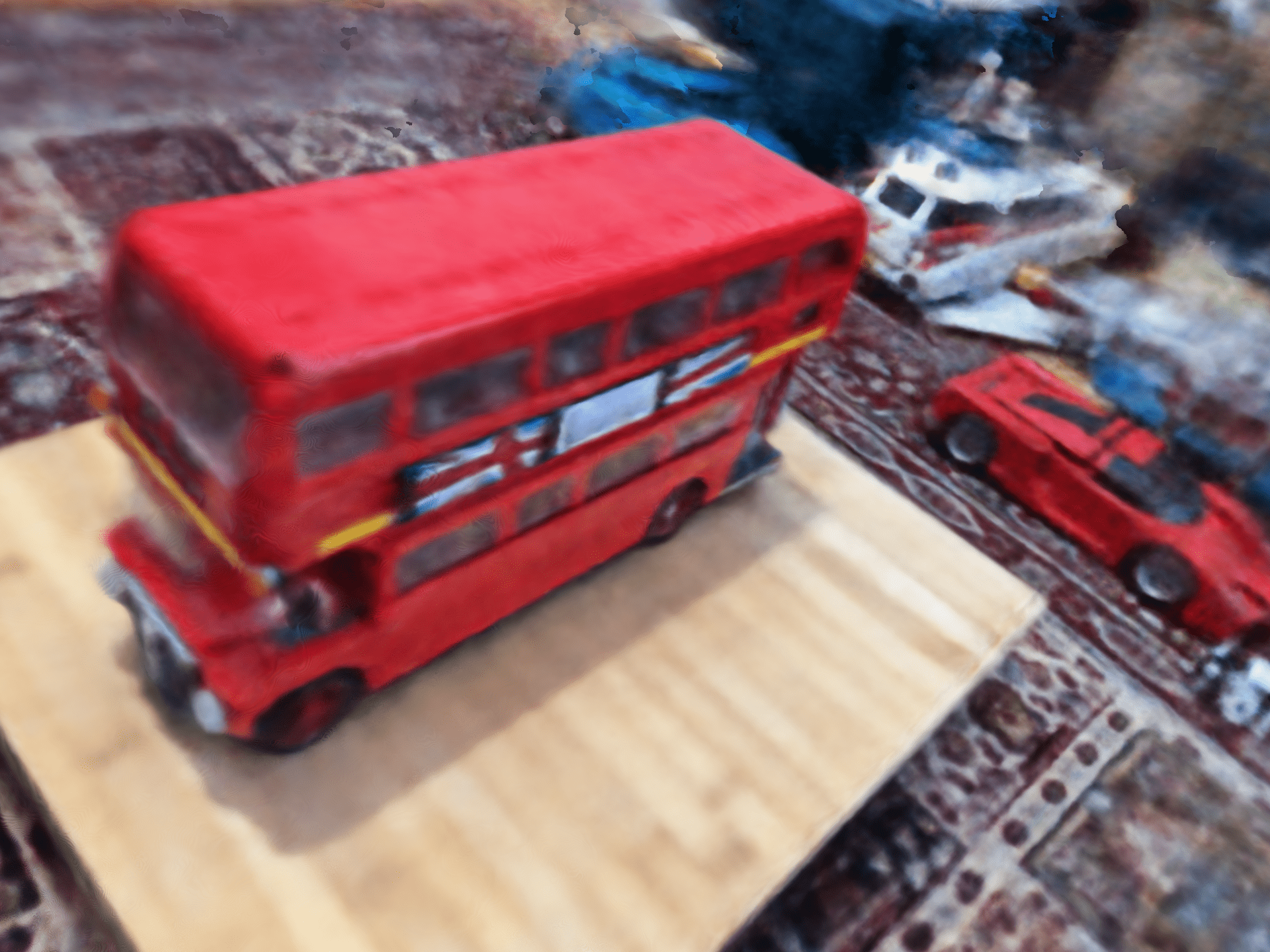}&
        \includegraphics[width=\linewidth]{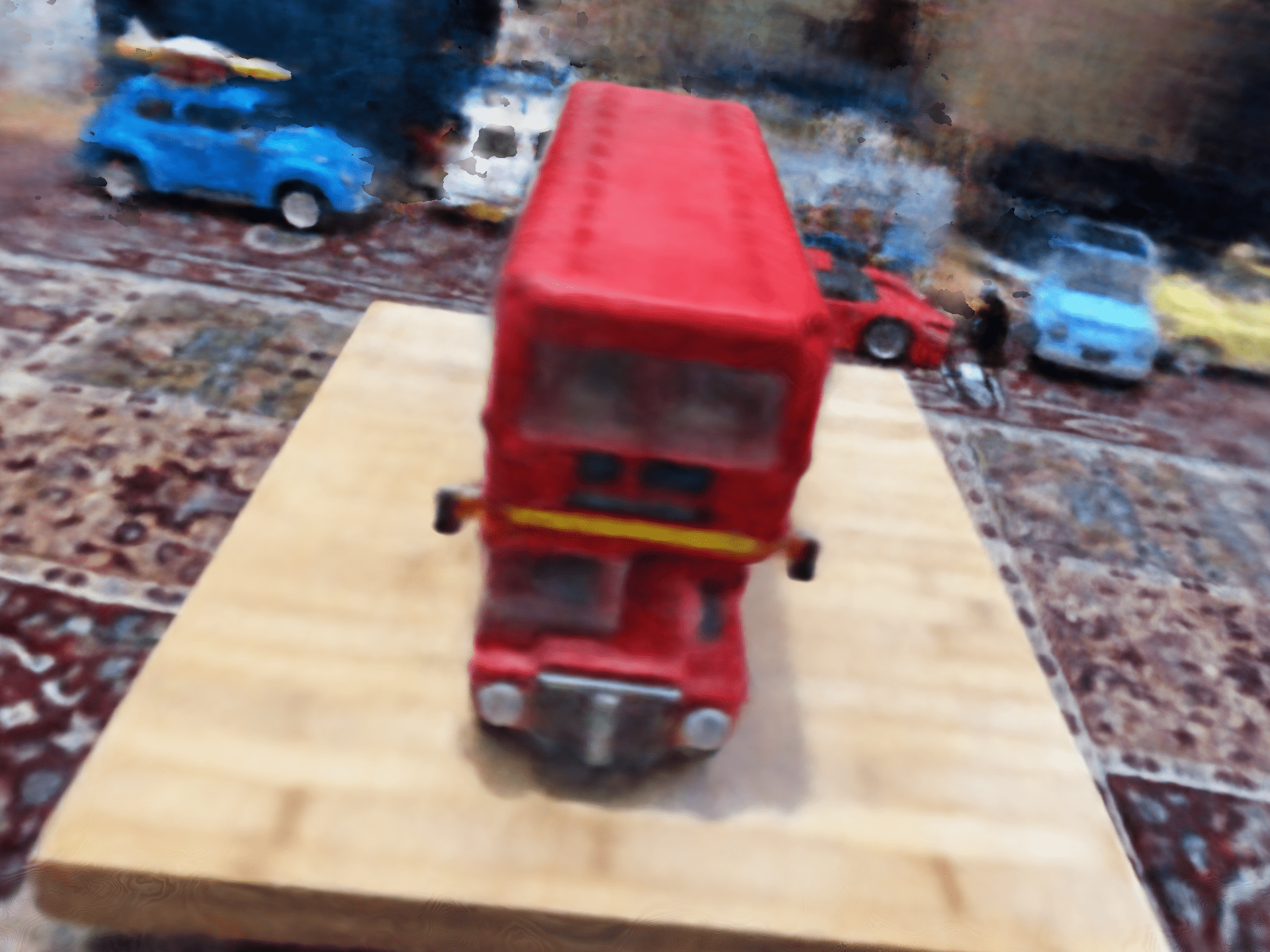} 
        \\
        NeuS &
        \includegraphics[width=\linewidth]{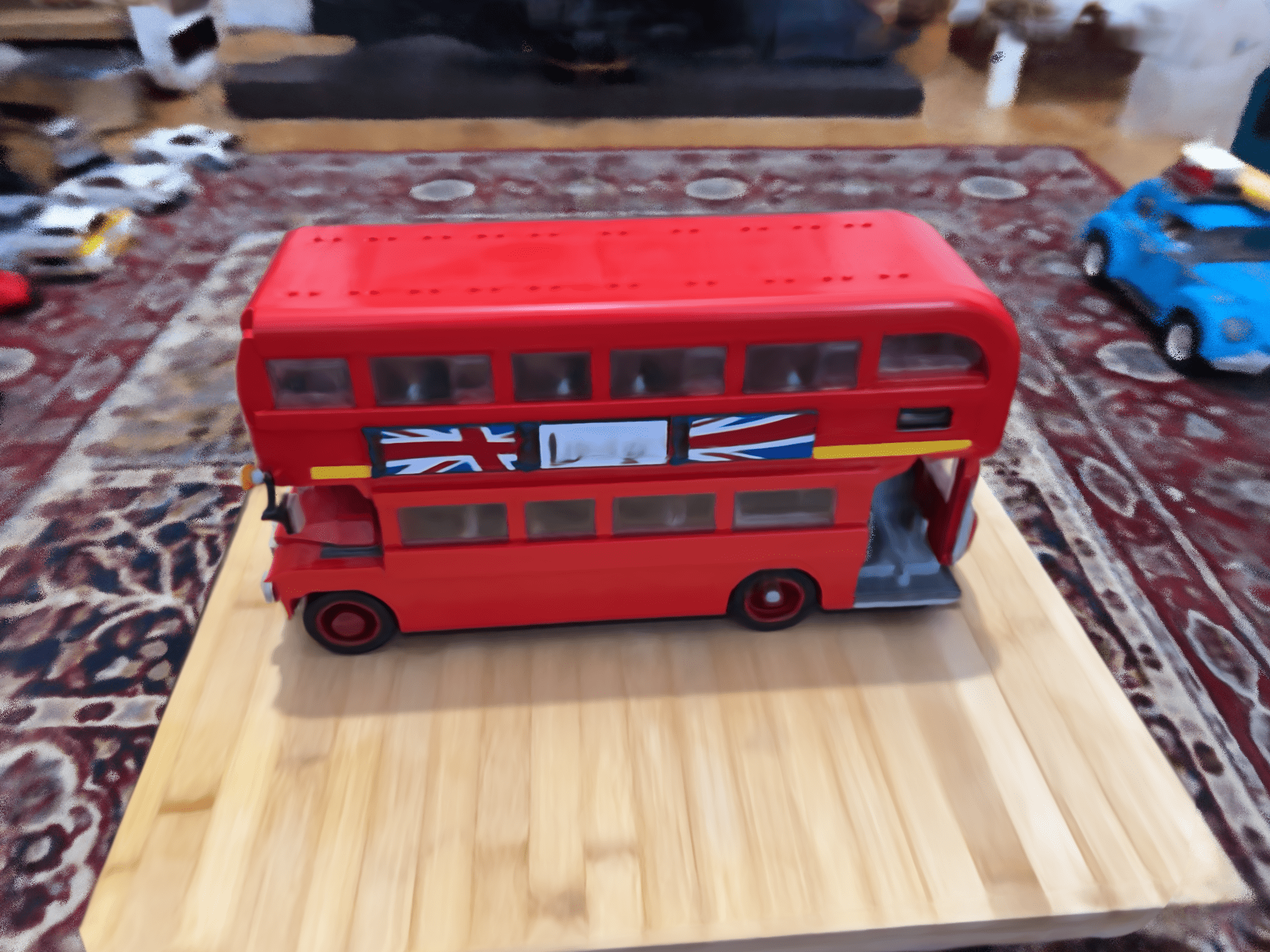}&
        \includegraphics[width=\linewidth]{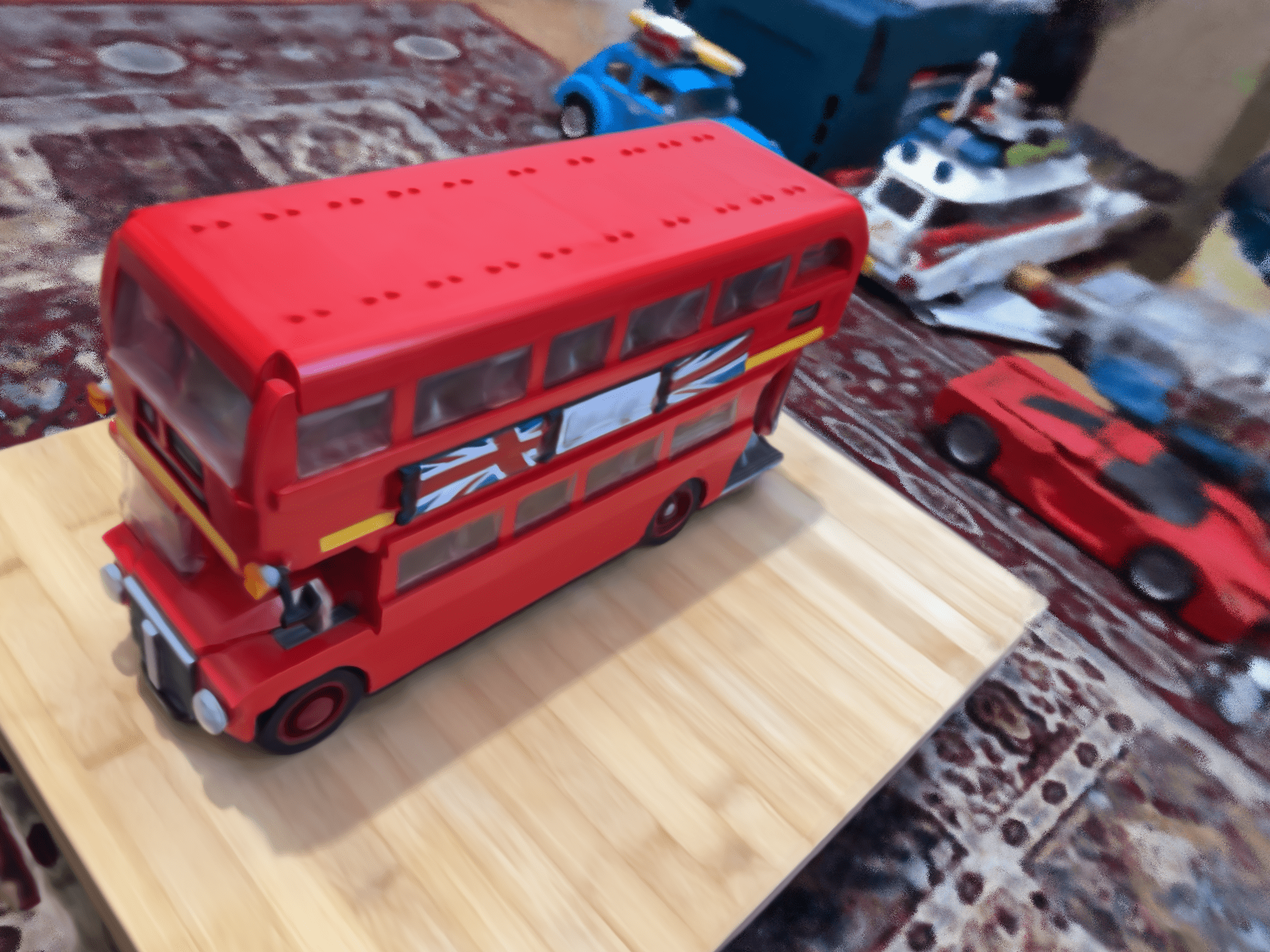}&
        \includegraphics[width=\linewidth]{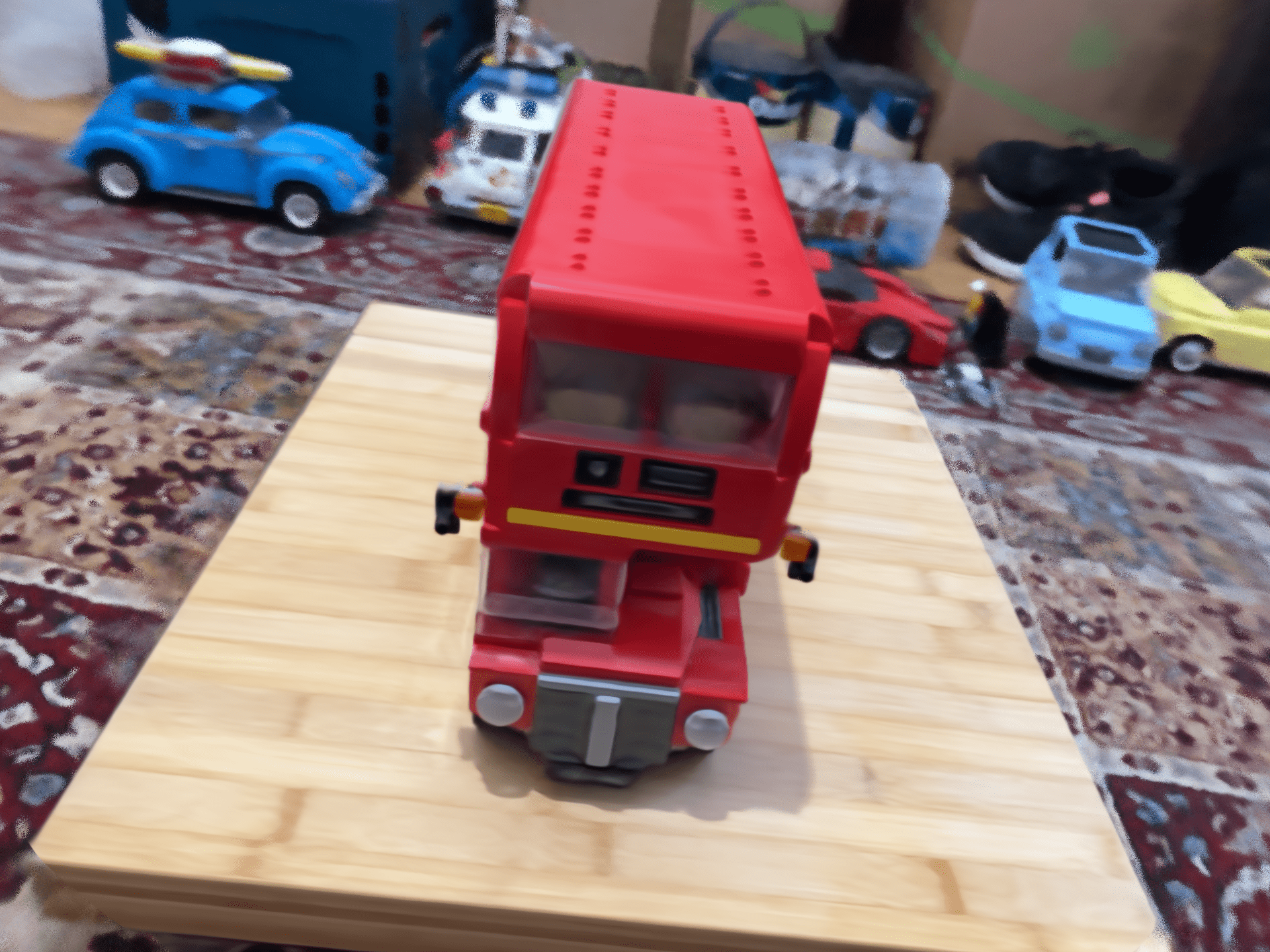} 
        \\
    \end{tabularx}
    
    \caption{Novel view synthesis visualisation.
\label{fig:nvs}}
\vspace{-0.3cm}
\end{figure}

\subsection{Preprocessing for Evaluation}\label{sec:preprocess}
The digital models cannot be used for evaluation directly because 1) they include the internal structures of LEGO bricks that are invisible after the bricks are connected, and 2) some surfaces are not visible during the image capture (see left in~\cref{fig:preprocess}). 
The evaluation for multi-view reconstruction should not expect a method to reconstruct invisible surfaces.
Therefore, it is necessary to compute a visibility mask of the 3D space so that only the visible regions are evaluated.

To compute the visibility mask and extract the external surfaces of a model, we place a virtual camera that shares the same extrinsic and intrinsic parameters at each frame and render a depth map of the aligned 3D model using the virtual camera.
Following DTU~\cite{jensen2014large}, there are two criteria we use to define the visible mask given a rendered depth map.
First, the space between the camera centre and the 3D points unprojected from the depth map should be visible. 
Second, we add a $+10mm$ offset to the depth values so that the visibility mask extends beyond the object surface, which sets the space slightly behind the object surface to be visible. 
This is necessary so that we can penalise incorrect reconstruction located behind the ground-truth object surface. 
The visibility mask of the entire space can be computed by taking the union of all visibility masks generated by all depth maps. 
A dense point cloud representing only the external surfaces of the 3D model for evaluating 3D reconstruction can be acquired by fusing all depth maps rendered from all viewpoints.

\subsection{Dataset Statistics}\label{sec:statistic}
The dataset comprises 153 models in total, 18 of which belong to the ``real-world model set'' and the rest are random models.
The number of LEGO bricks used in the models varies significantly.
Models in the ``real-world model set'', some of which take thousands of bricks to build, are generally more complex than the ``random model set''. 
To ensure models in the ``random model set'' have diverse geometry, we only use random models with at least $25$ bricks.
Although the first two stages of the annotation pipeline require labour effort for the keypoint annotation, and pose refinement, with the help of the GUI, it takes about $10$ minutes per sequence.  

\begin{table}[t]
    \centering
    \begin{tabular}{c c c c}
    Methods & PSNR$\uparrow$ & SSIM$\uparrow$ & LPIPS$\downarrow$  \\
    \hline
    \hline  
    NeRF~\cite{mildenhall2020nerf} & 20.66 & 0.64 & 0.49 \\
    DVGO~\cite{sun2022direct} & 20.64 & \textbf{0.67} & \textbf{0.44} \\
    NeuS~\cite{wang2021neus} & \textbf{21.49} & \textbf{0.67} & 0.48 \\
    \hline
    \hline  
    \end{tabular}
    \caption{Novel view synthesis experiment. NeuS and NeRF outperform DVGO by a large margin.}
    \label{tab:nvs_exp}
\end{table}
\begin{table*}[t]
    \centering
    \begin{tabular}{c c c c c c c c}
    Methods & Abs Rel $\downarrow$ & Sq Rel $\downarrow$ & RMSE $\downarrow$ & RMSE log $\downarrow$ & $\delta \textless 1.25 \uparrow$ & $\delta \textless 1.25^2 \uparrow$ & $\delta \textless 1.25^3 \uparrow$\\
    \hline
    \hline  
    NN-interpolation & 0.0845 & 0.0177 & 0.0663 & 0.1327 & 0.9360 & 0.9748 & 0.9878 \\
    MSPF (baseline)& 0.0806 & 0.0144 & 0.0588 & 0.1231 & 0.9413 & 0.9787 & 0.9902 \\
    \makecell{MSPF (finetuned)} & \textbf{0.0678} & \textbf{0.0056} & \textbf{0.0413} & \textbf{0.0920} & \textbf{0.9441} & \textbf{0.9932} & \textbf{0.9983} \\
    \hline
    \hline  
    \end{tabular}
    \caption{Depth map enhancement experiment. The improvement brought by finetuning on our dataset is larger than the improvement from NN-interpolation to the MSPF baseline in most of the evaluation metrics.}
    \label{tab:depth_exp}
\end{table*}
\section{Experiments}
We evaluate three tasks using the proposed dataset: 1) multi-view surface reconstruction; 2) novel view synthesis; 3) and colour-guided depth map enhancement. 
$18$ models in the real-world model set are used for evaluation, and models and images in the random model set are used for training.
To show that the random model set can generalise to the real-world model set when evaluating methods that require training data, we report results of two versions: one that is provided by the authors and pretrained on external datasets, and one that is finetuned on the random model set.

\subsection{Evaluating Dataset Annotations}
\begin{table}[t]
    \centering
    \small
    \begin{tabular}{c c c c c}
    & \makecell{RMSE \\ (mm)} $\downarrow$ & \makecell{rotation diff. \\(in \degree)} $\downarrow$  & \makecell{Acc.\\(1mm)} $\uparrow$
    & \makecell{Rec.\\(1mm)} $\uparrow$
    \\
    \hline
    \hline
    ARKit & 4.454 & 0.581 & 90.1\% & 91.3\%\\
    Ours & \textbf{2.060} & \textbf{0.522} & \textbf{93.7\%} & \textbf{93.8}
    \\
    \hline
    \hline
    \end{tabular}
    \caption{Pose evaluation and shape deviation caused by imperfect poses on calibration sequences.}
    \label{tab:pose}
    \vspace{-0.2cm}
\end{table}

\begin{table}[t]
    \centering
    \small
    \begin{tabular}{c c c}
    Acc.(1mm) $\uparrow$ & Rec.(1mm) $\uparrow$ \\
    \hline
    \hline
    85.9\% & 84.0\%\\
    \hline
    \hline
    \end{tabular}
    \caption{Impact of perturbed poses on shape reconstruction on synthetic data.}
    \label{tab:syn_pose}
    \vspace{-0.5cm}
\end{table}

We provide analyses on the quality of the ground-truth shape annotations and camera alignment.
Despite the ground-truth shapes theoretically are the replicas of the actual objects used for image capture, shape errors still exist in the model construction process.
To investigate the effect of accumulated error of model building, we measure the 3D dimensions of $20$ models randomly selected from the dataset using a micrometre with an accuracy of $0.05mm$. 
The difference in average is $0.2 mm$ ($200.5 \mu m$). 

We also evaluate quantitatively the accuracy of camera alignment.
Since the exact ground-truth camera poses are impossible to attain for sequences captured on a mobile phone, we use camera poses obtained from the camera calibration algorithm as ground-truth poses to compare with the poses given by the proposed alignment algorithm ARKit respectively.
Specifically, we record $10$ calibration sequences, where a randomly selected LEGO model is put on a ChArUco calibration board, which is robust to occlusions caused by the LEGO model.
We then use the camera calibration algorithm in the OpenCV library~\cite{opencv_library} to compute the reference camera poses for evaluation.

We report the pairwise Root Mean Square Error for translation and degree difference for rotation in~\cref{tab:pose}. 
The proposed algorithm reduces the errors in translation and rotation by almost $100\%$ and $12\%$ respectively.
However, the metrics on camera poses do not reveal how the 3D reconstruction would be affected by the imperfect camera poses.
Therefore, we run COLMAP for each sequence to reconstruct the 3D shapes using three sets of camera poses: 1) the reference camera poses (from camera calibration); 2) the refined poses proposed in the paper; and 3) the poses from ARKit. 
We then compare the shapes reconstructed using the refined poses and ARKit poses against the ones using the reference poses, such that we can quantify the effect of imperfect poses on 3D reconstruction. 
The results are reported in ~\cref{tab:pose}, showing that 
the proposed algorithm outperforms ARKit in metrics for 3D reconstruction.

We further evaluate on synthetic data to eliminate the potential bias on the calibration sequences as the presence of the calibration board might improve the camera tracking.
We first render ground-truth depth maps using ground-truth poses for each LEGO model. 
The ground-truth poses are then perturbed based on the publicly available error statistics of ARKit tracking~\cite{kim2022benchmark}. 
We then run TSDF-Fusion using the GT depth maps with these two sets of poses and compare the reconstructions, as reported in ~\cref{tab:syn_pose}.
Similar to the results on the real calibration sequences, most of the points are within $1mm$, which is well below the thresholds used in the experiments.

\subsection{Multi-view Reconstruction}
\boldstart{Setup} We investigate the performance of a wide range of multi-view reconstruction algorithms with different types of input data. 
We take the state-of-the-art approach in each category in the evaluation.
Specifically, COLMAP~\cite{schoenberger2016sfm}, a popular framework in the community for Multi-view Stereo (MVS), is chosen as a representative for methods based on traditional multi-view geometry. 
NeuS~\cite{wang2021neus} and NeRF~\cite{mildenhall2020nerf} are selected to represent the neural field based approaches.  
Vis-MVSNet~\cite{zhang2020visibility} is a representative of deep-learning-based MVS solutions.
In addition to these RGB-only methods, we also include an RGBD approach, RGBD neural surface reconstruction~\cite{azinovic2022neural}, and two depth-only fusion algorithms, TSDF-Fusion~\cite{zhou2018open3d}, and BNV-Fusion~\cite{li2022bnv}, to investigate the effect of low-resolution depth maps captured by a mobile device in multi-view reconstruction.
Among all methods in this experiment, only Vis-MVSNet requires a training stage. 
Therefore, we report the results of two versions of Vis-MVSNet, one that is pretrained on BlendedMVS~\cite{yao2020blendedmvs}, a popular MVS benchmark, provided by the author, and the other one that is finetuned on the random model set.
Following common practice in MVS, we use Chamfer Distance, precision, recall, and F1 score as evaluation metrics.
We set two thresholds ($2.5mm$, $5mm$) for precision and recall.

\begin{figure}[th]
    \centering
    \newcolumntype{Y}{>{\centering\arraybackslash}X}
    \begin{tabularx}{\linewidth}{@{}X@{}Y@{\,}Y@{\,}Y@{}}
        RGB &
        \includegraphics[width=\linewidth,height=1.8cm,keepaspectratio,origin=c,angle=270]{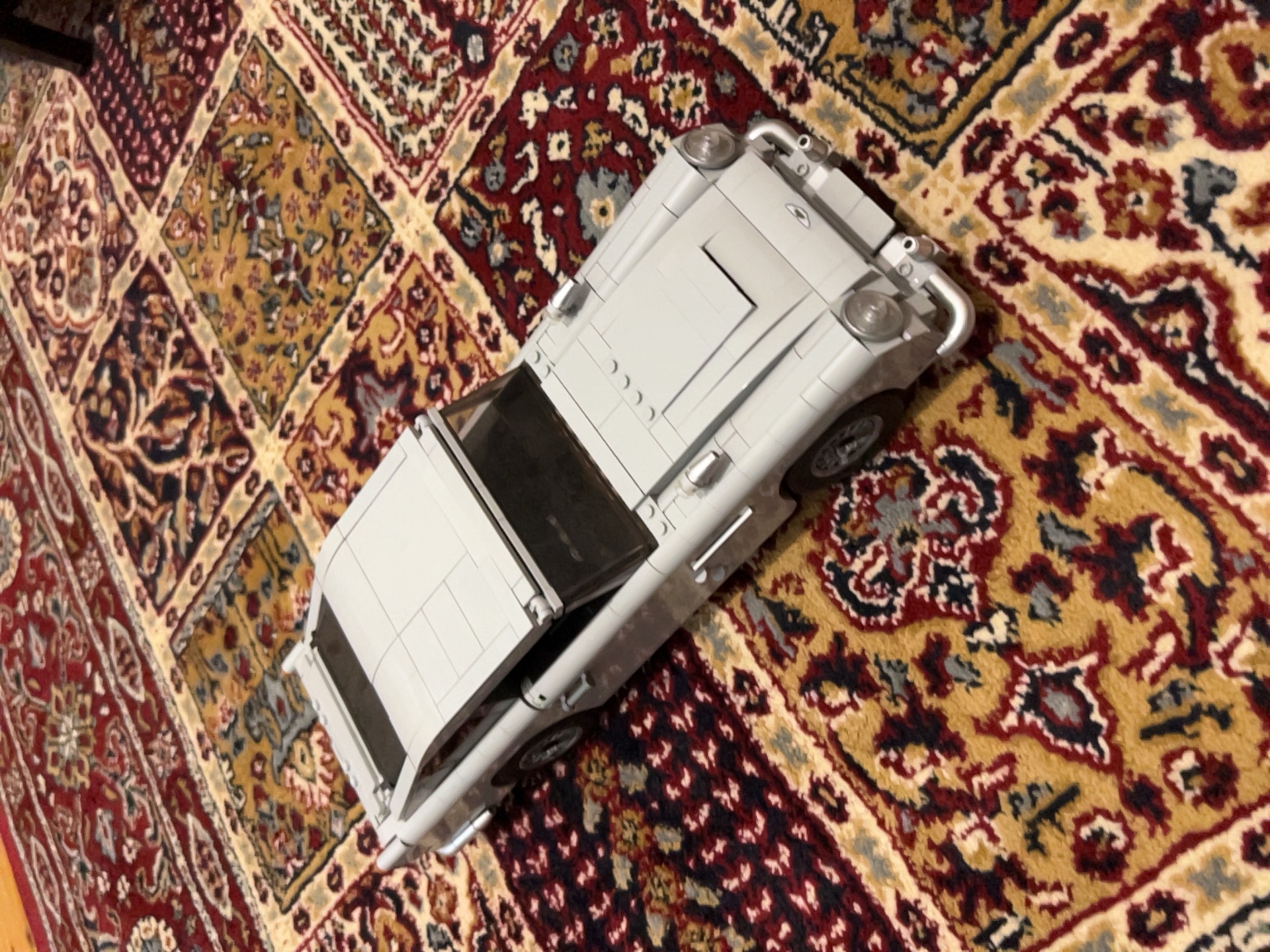}&
        \includegraphics[width=\linewidth,height=1.8cm,keepaspectratio,origin=c,angle=270]{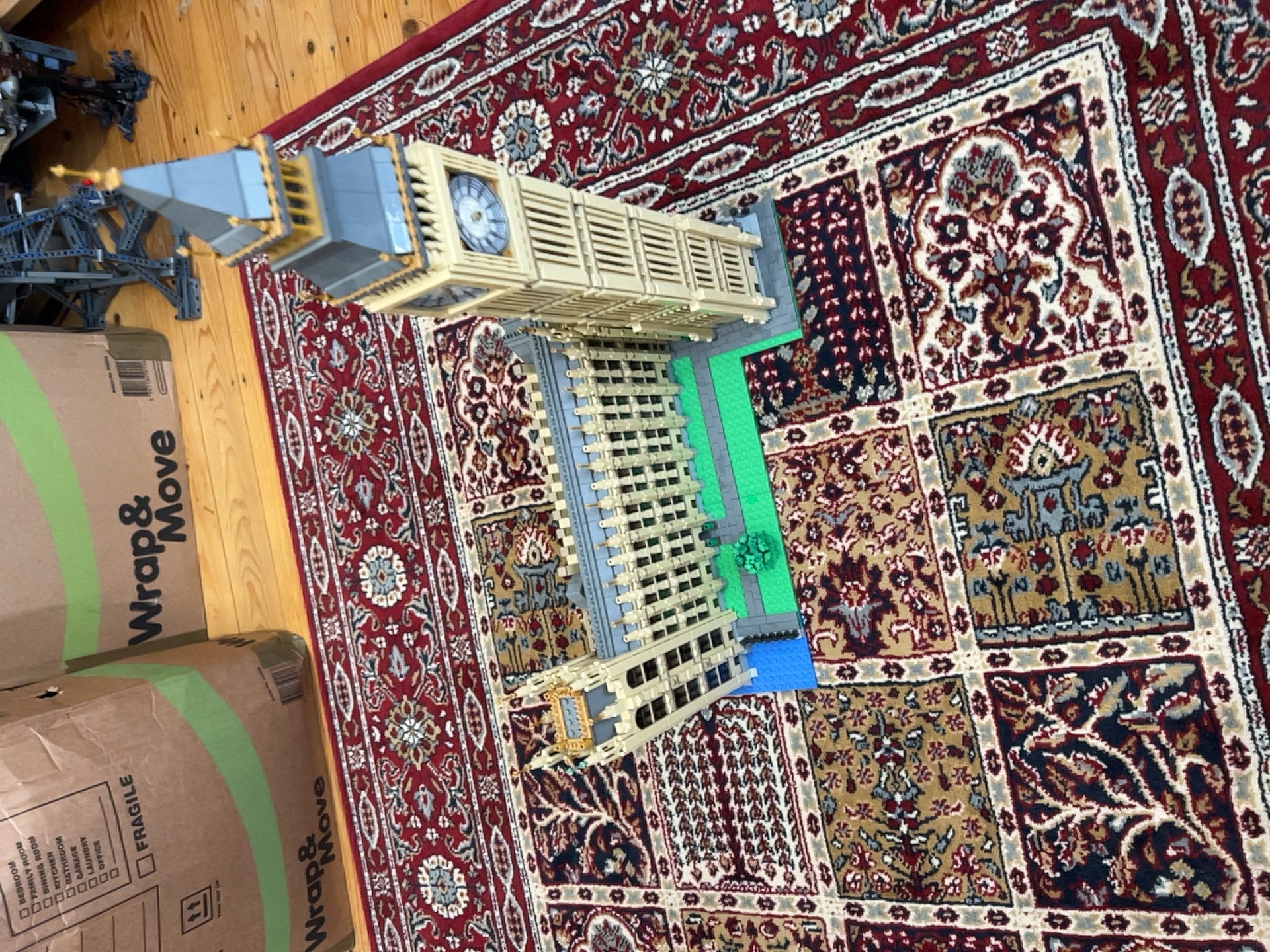}&
        \includegraphics[width=\linewidth,height=1.8cm,keepaspectratio,origin=c,angle=270]{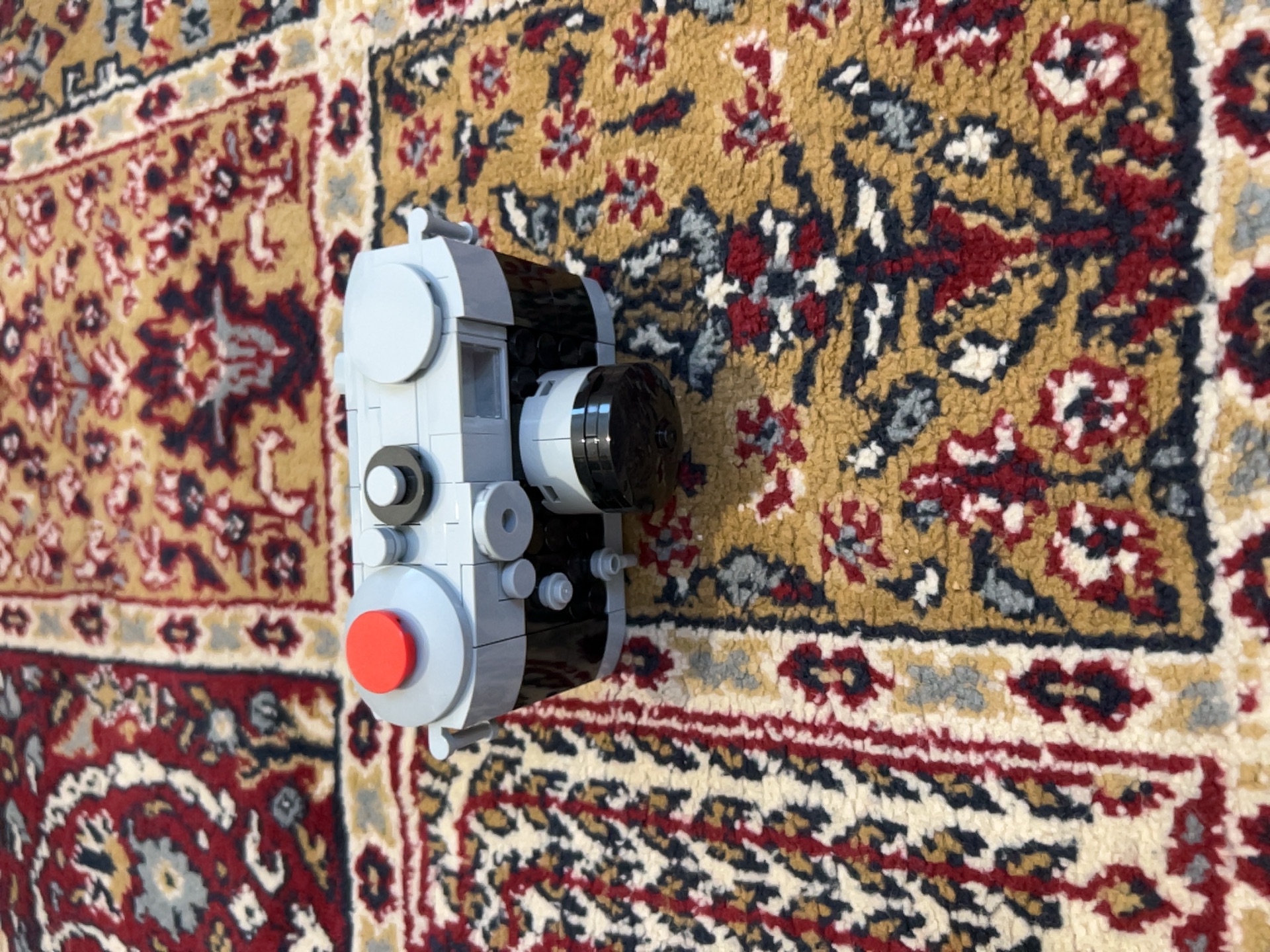}
        \\
        TSDF-Fusion &
        \includegraphics[width=\linewidth,height=2cm,keepaspectratio]{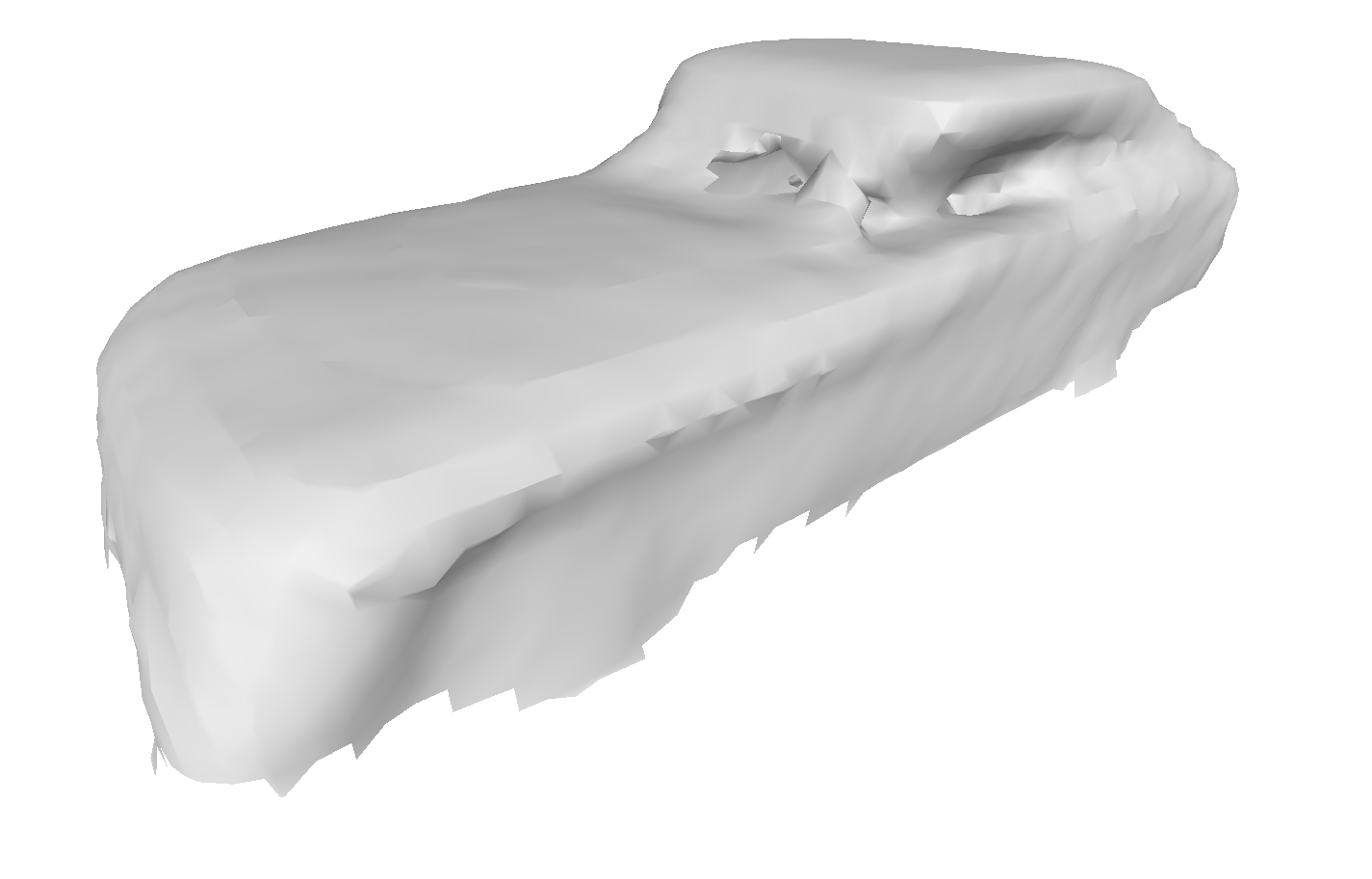}&
        \includegraphics[width=\linewidth,height=2cm,keepaspectratio]{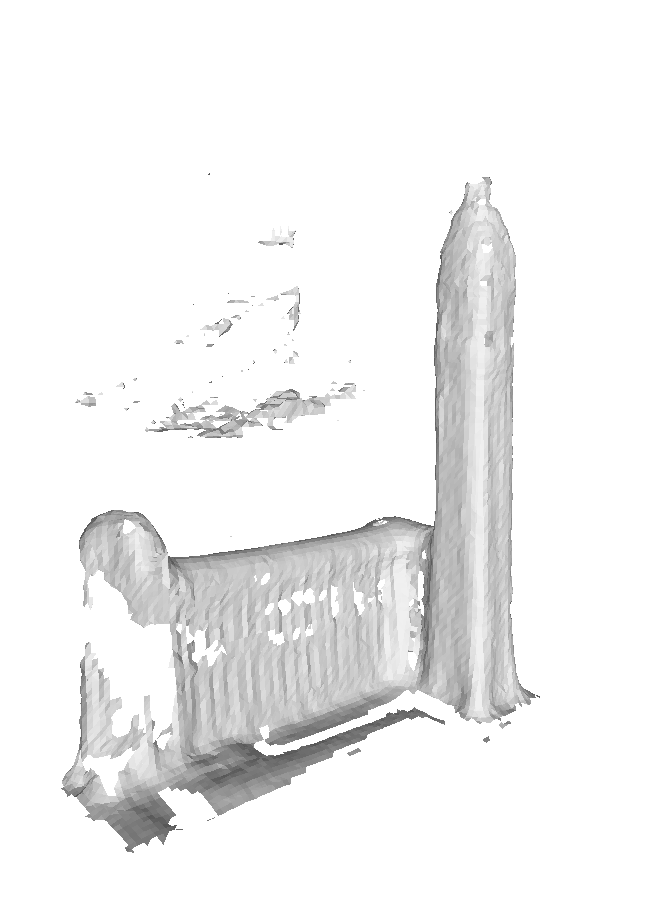}&
        \includegraphics[width=\linewidth]{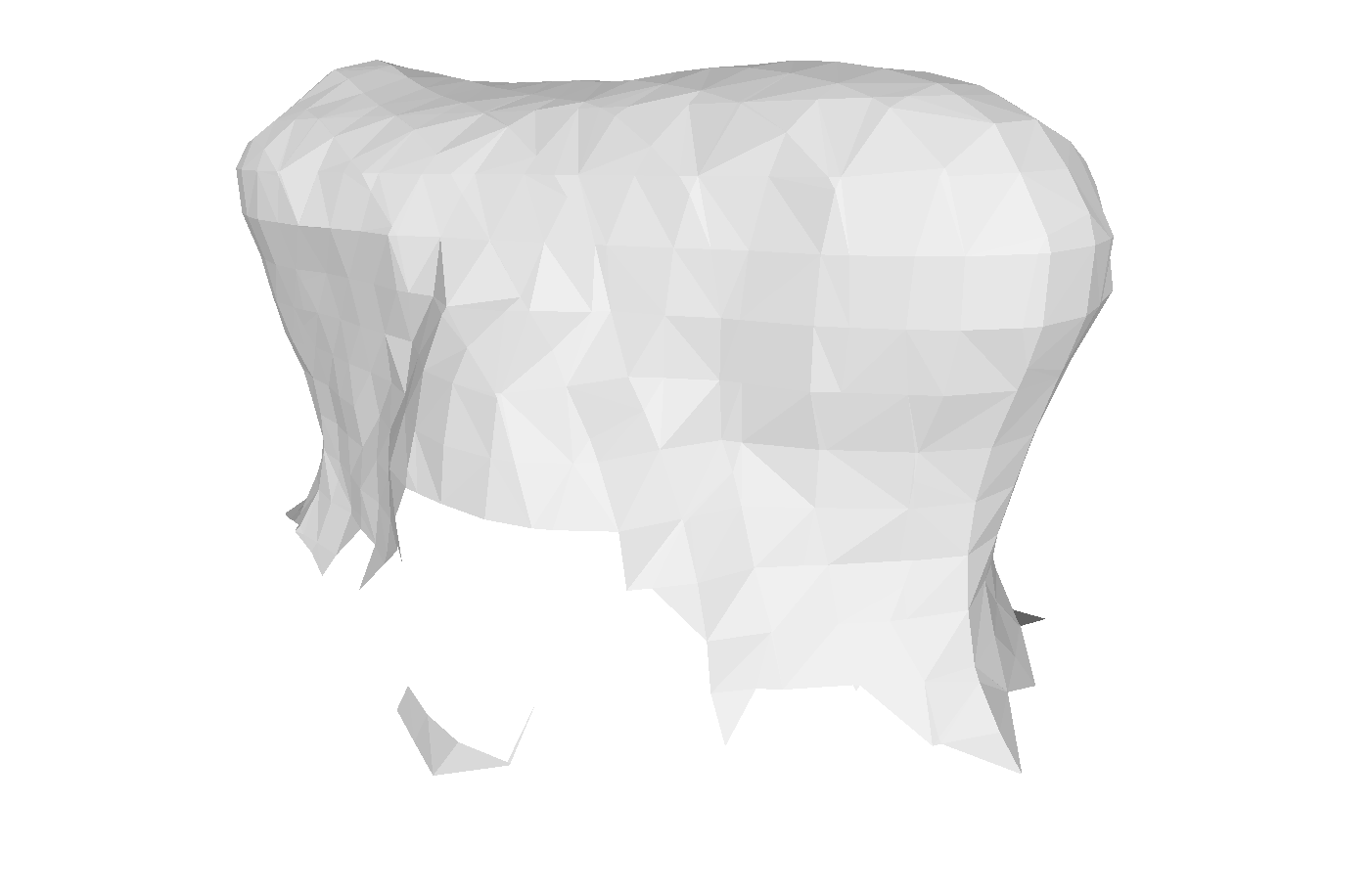}
        \\
        BNV-Fusion &
        \includegraphics[width=\linewidth]{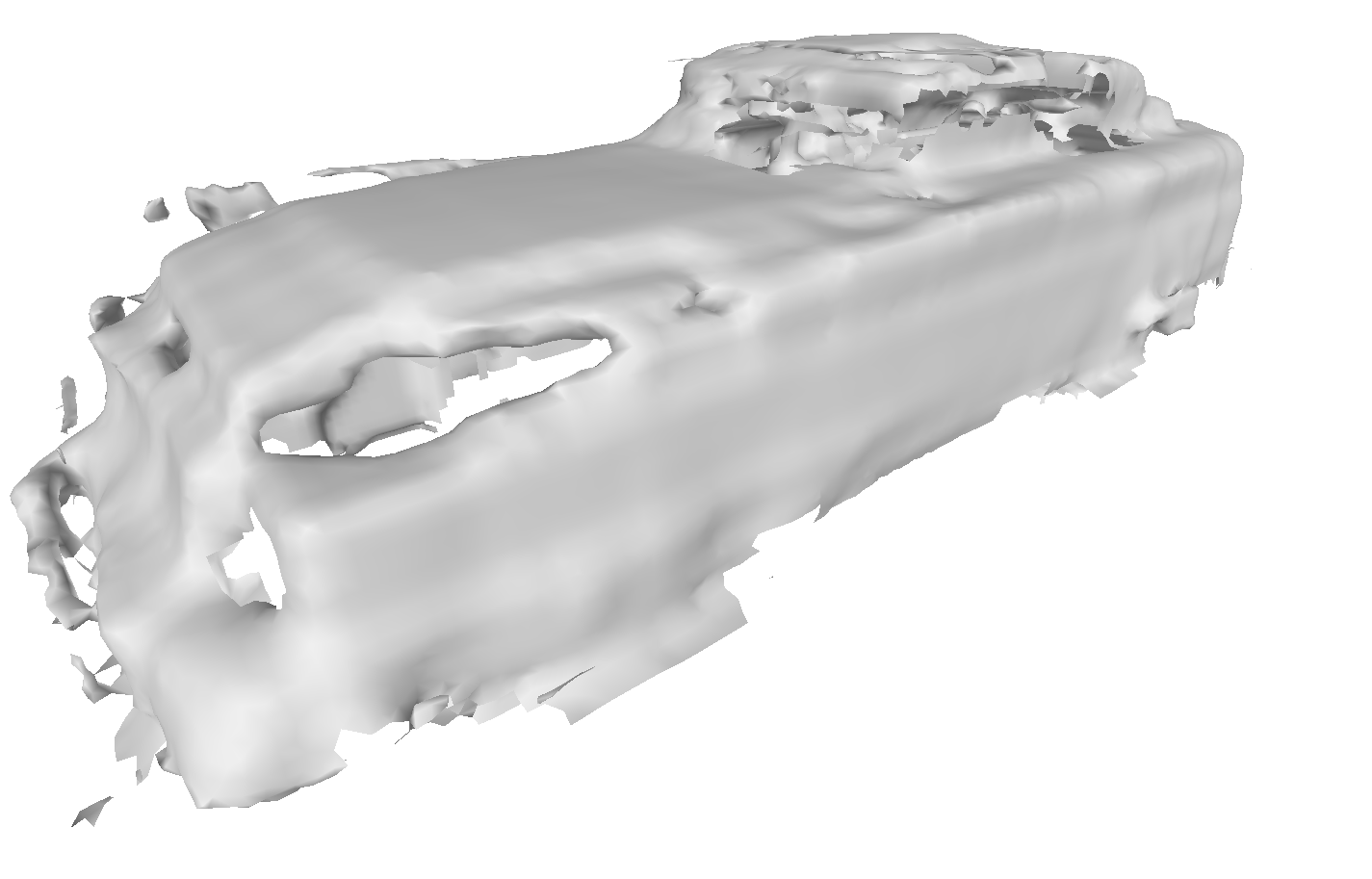}&
        \includegraphics[width=\linewidth,height=2cm,keepaspectratio]{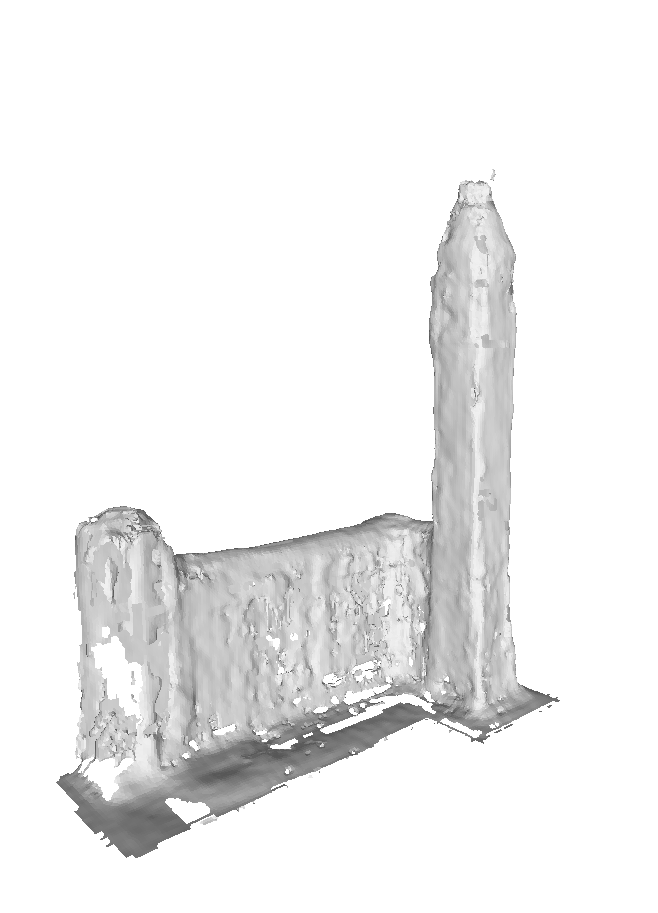}&
        \includegraphics[width=\linewidth]{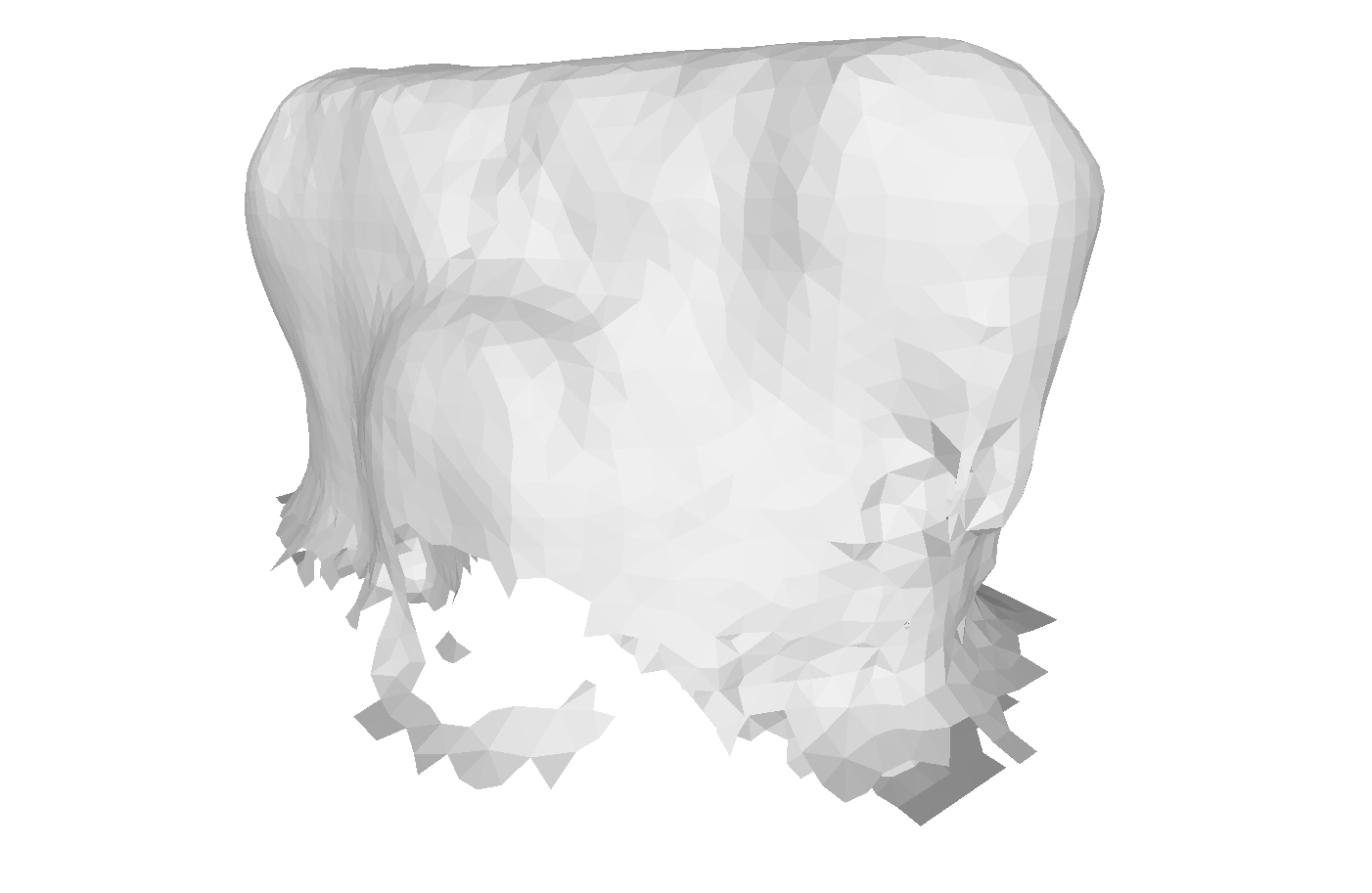}
        \\
        Neural-RGBD &
        \includegraphics[width=\linewidth]{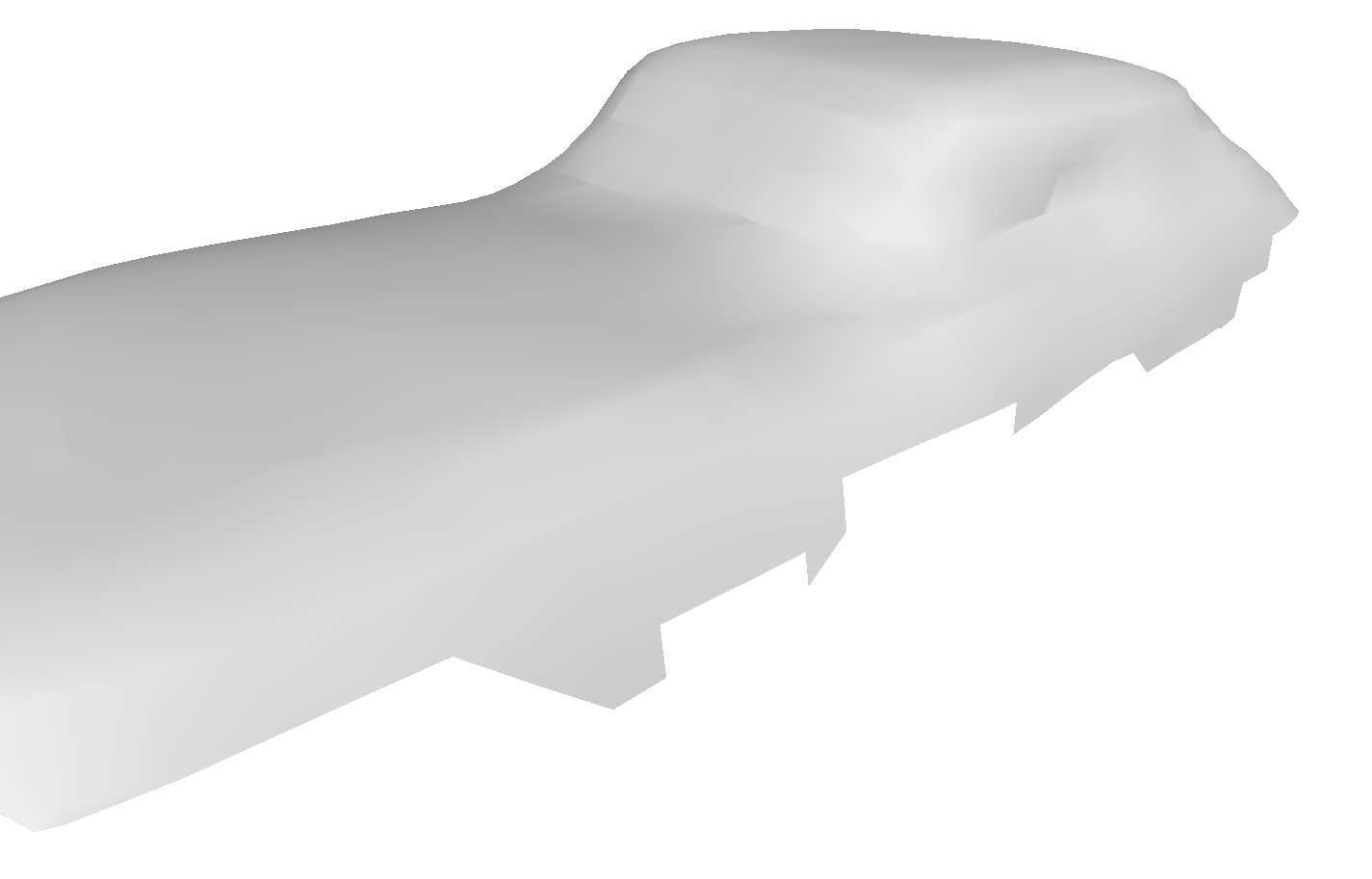}&
        \includegraphics[width=\linewidth,height=2cm,keepaspectratio]{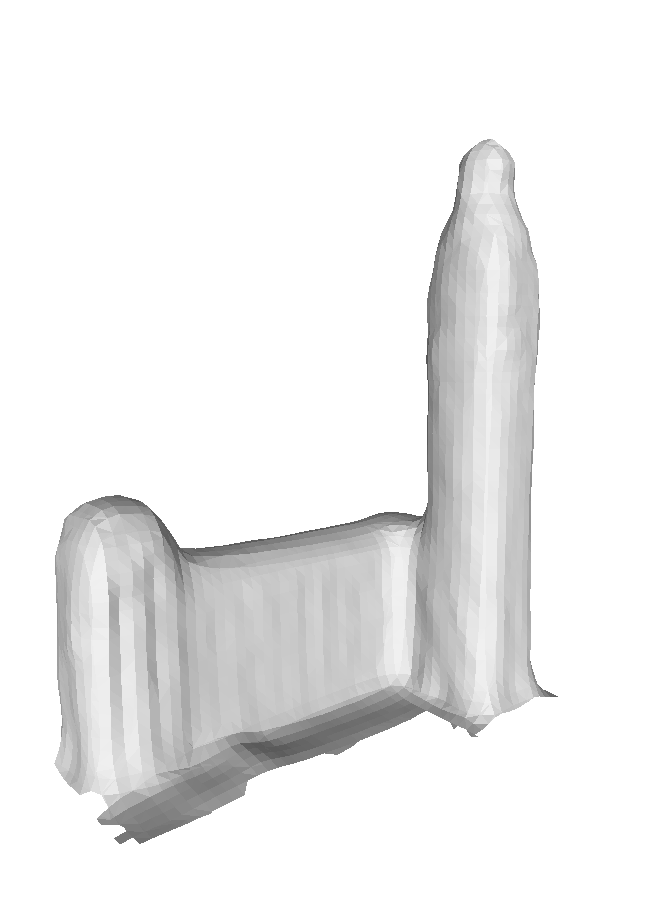}&
        \includegraphics[width=\linewidth]{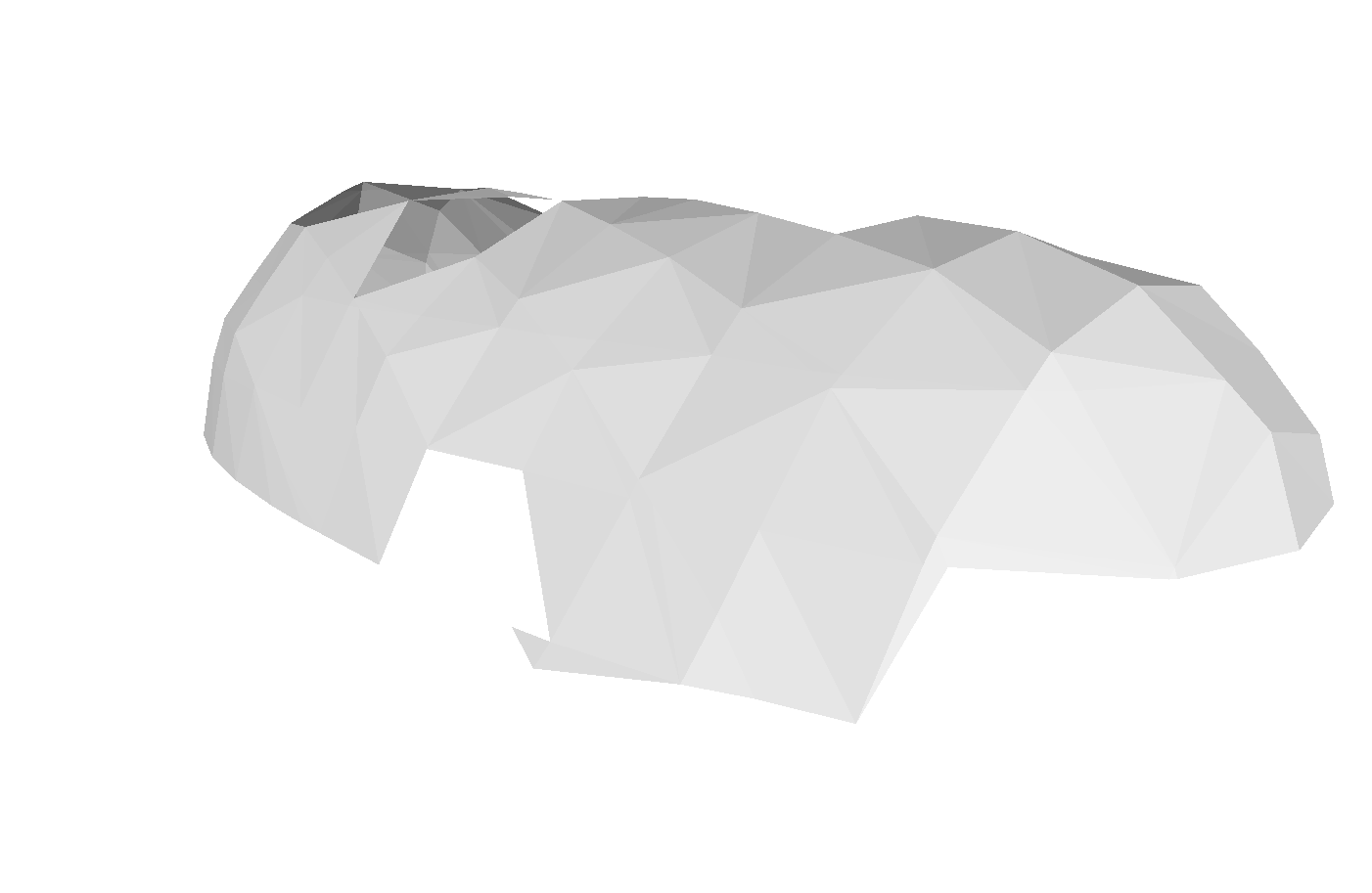}
        \\
        COLMAP &
        \includegraphics[width=\linewidth]{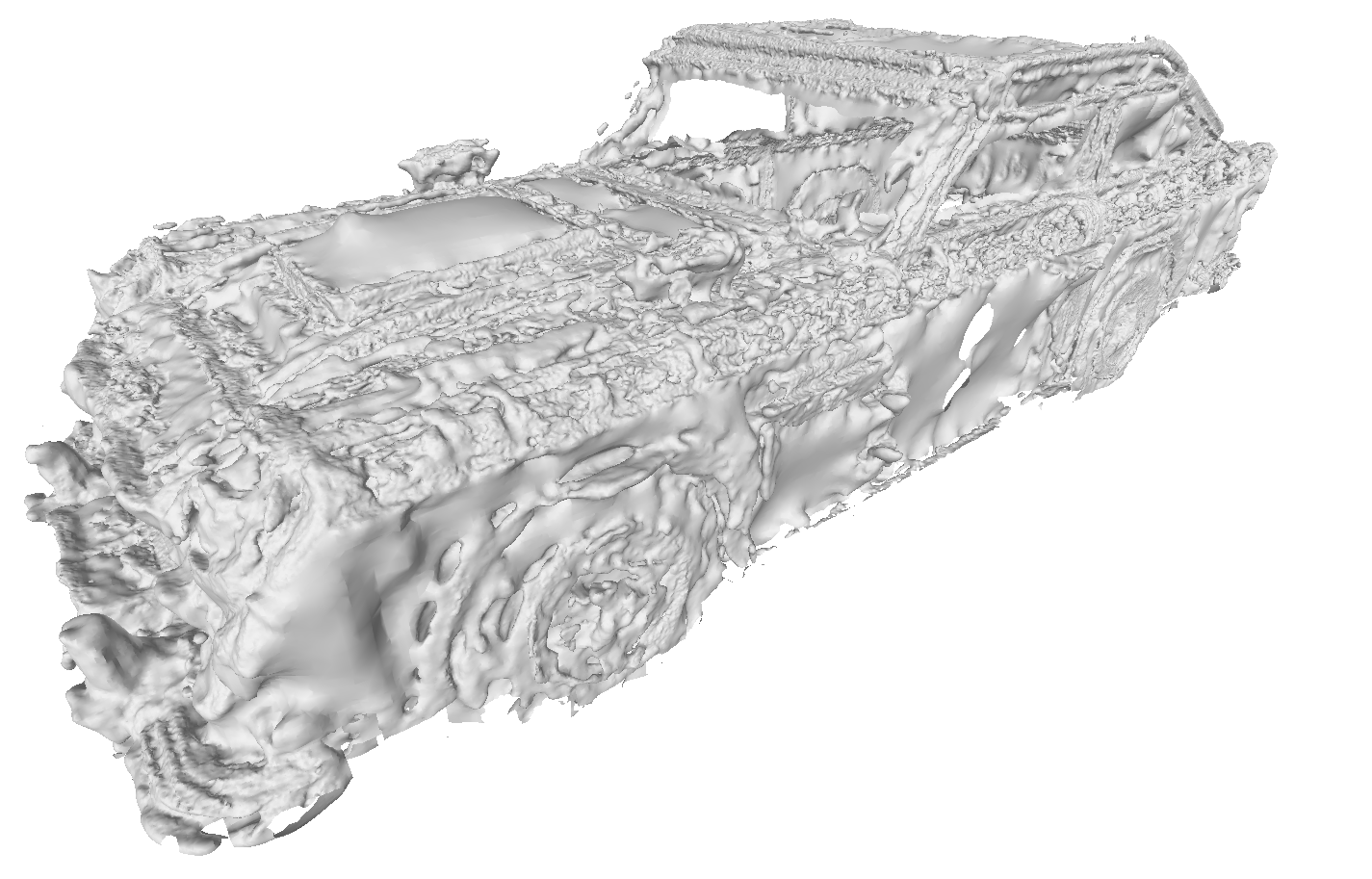}&
        \includegraphics[width=\linewidth,height=2cm,keepaspectratio]{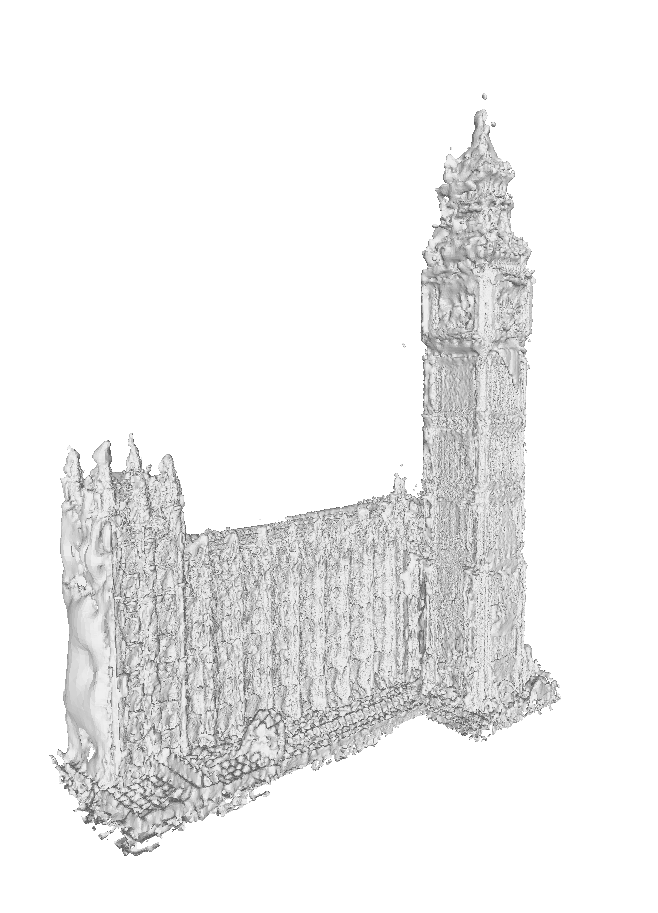}&
        \includegraphics[width=\linewidth]{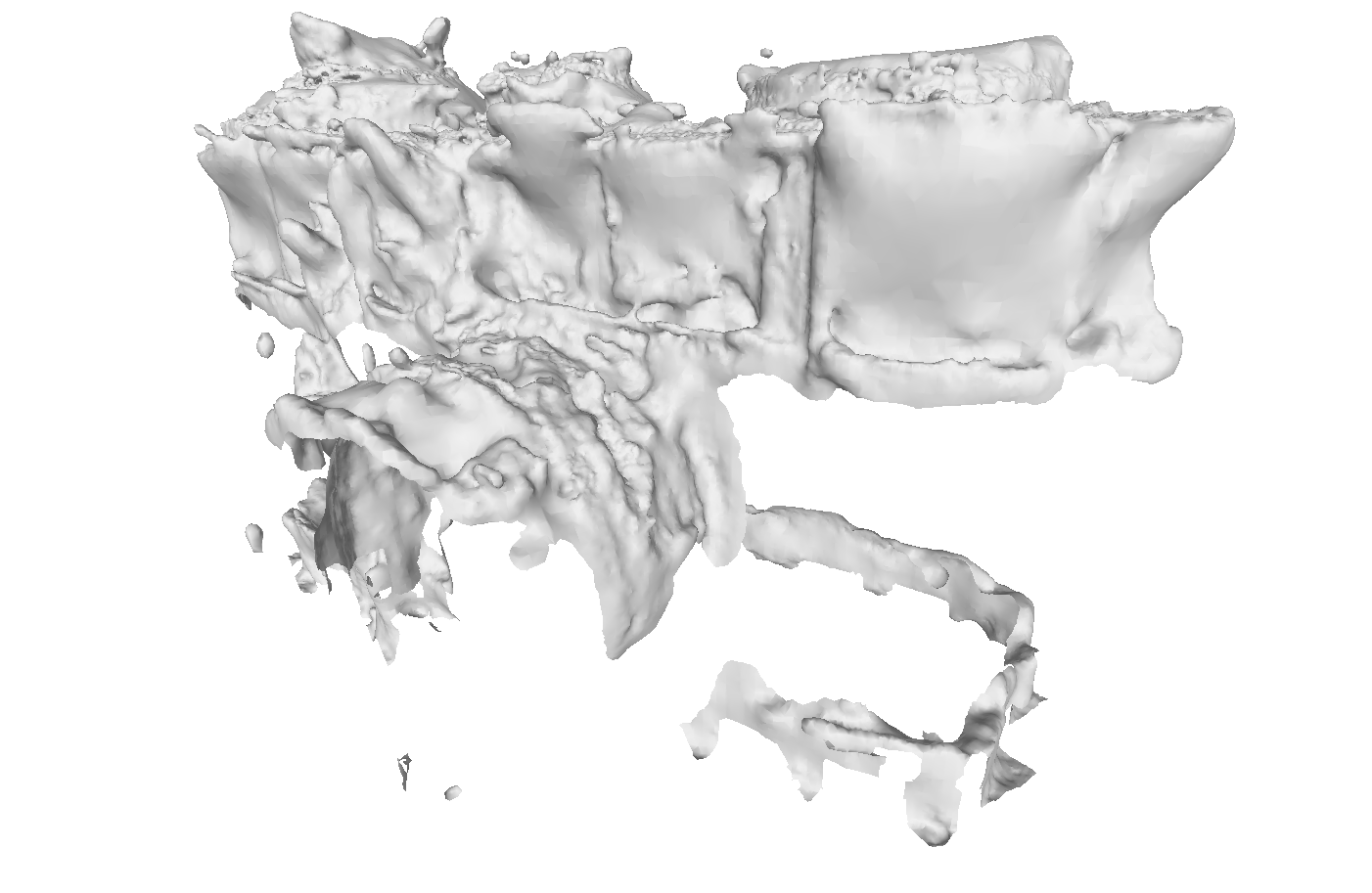}
        \\
        Vis-MVSNet &
        \includegraphics[width=\linewidth]{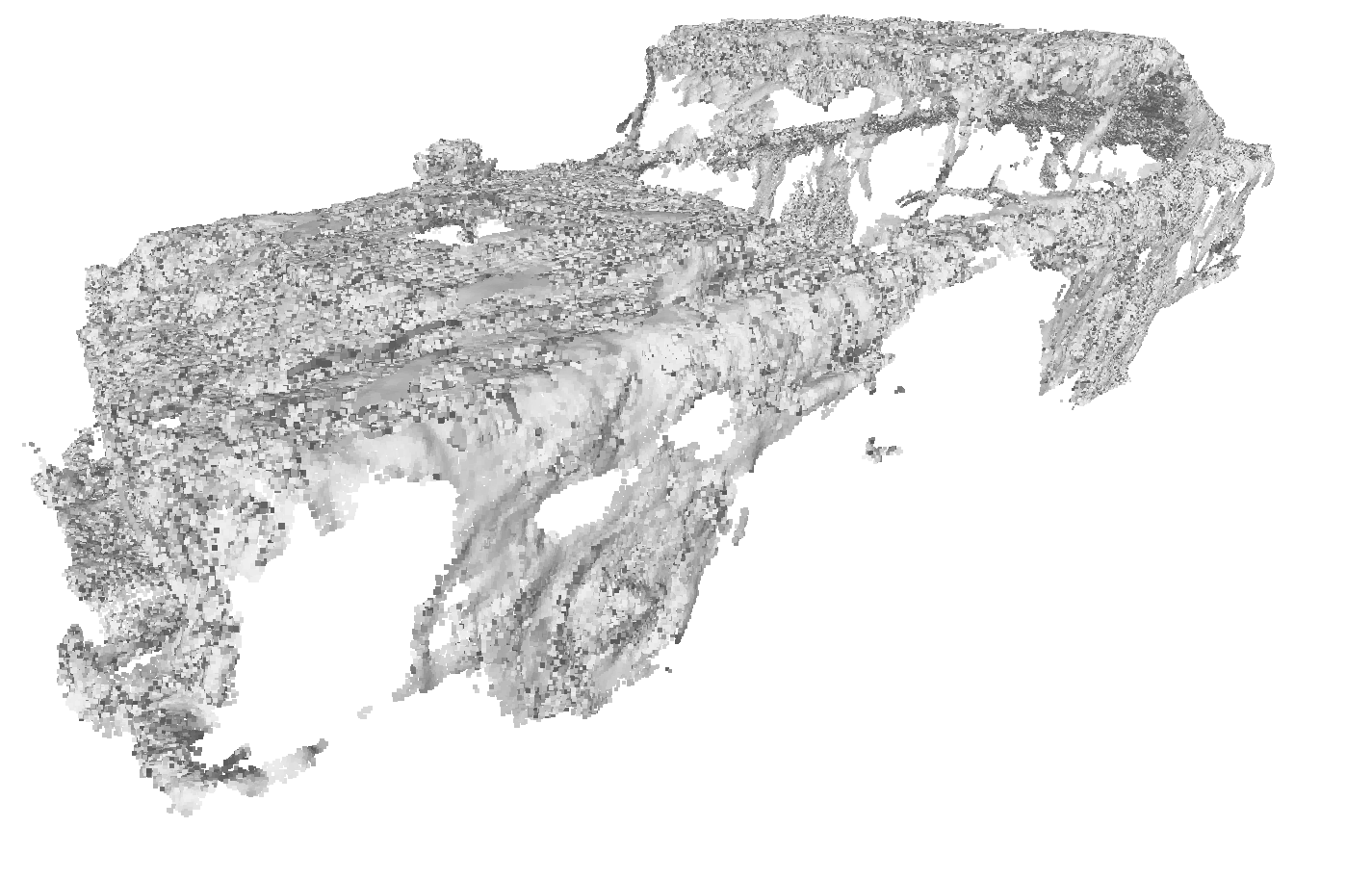}&
        \includegraphics[width=\linewidth,height=2cm,keepaspectratio]{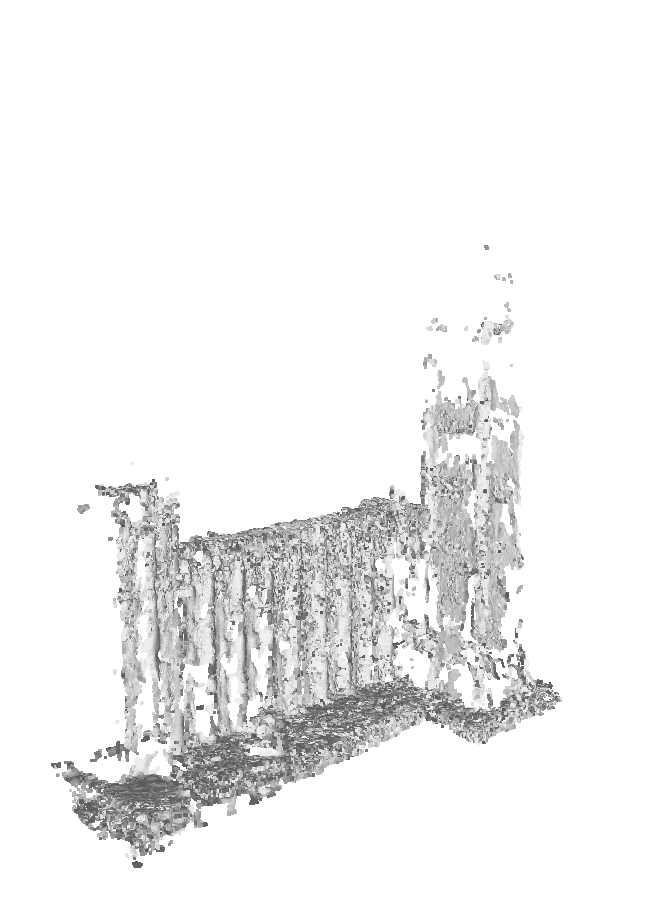}&
        \includegraphics[width=\linewidth]{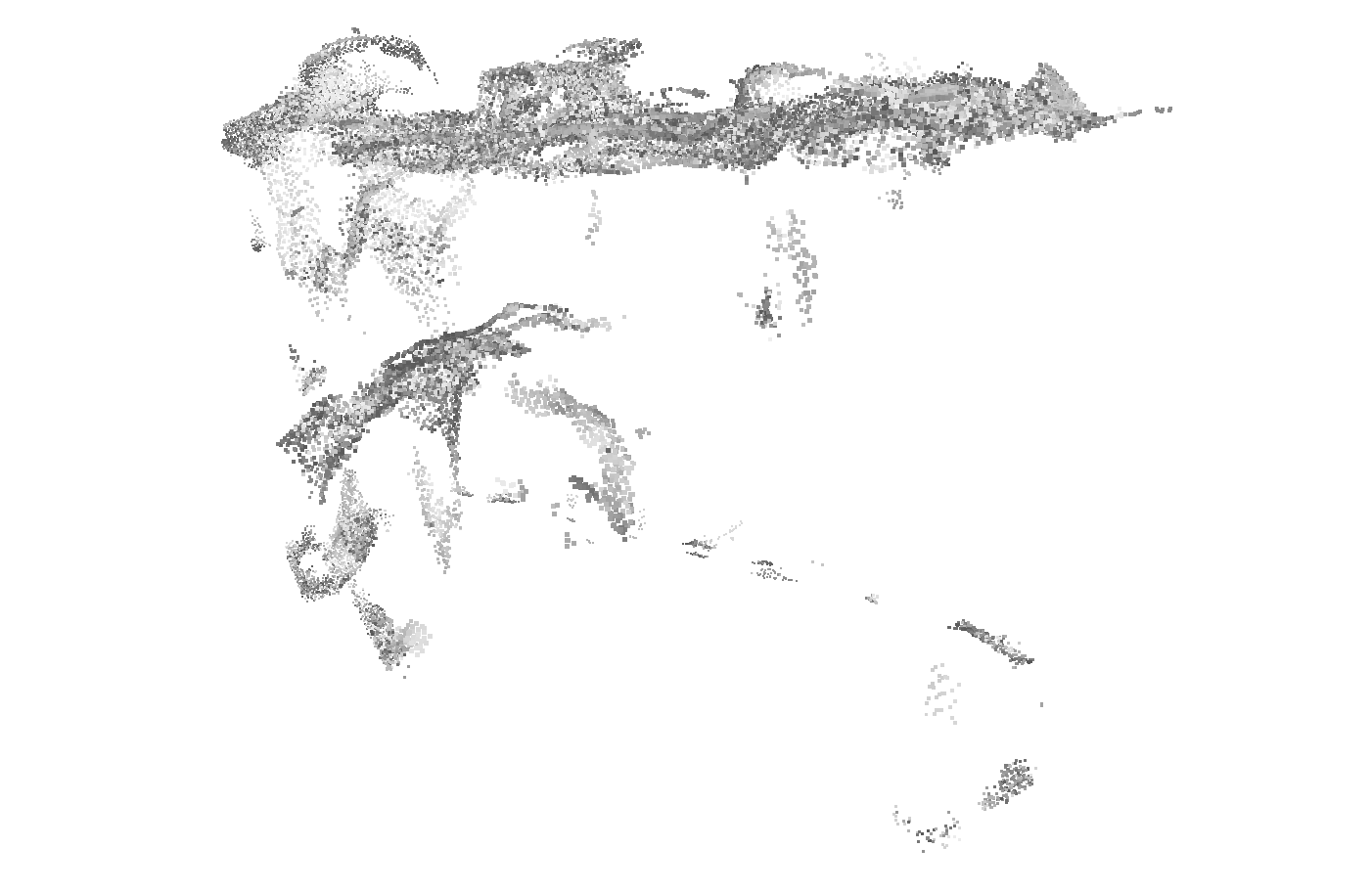}
        \\
        Vis-MVSNet (finetuned) &
        \includegraphics[width=\linewidth]{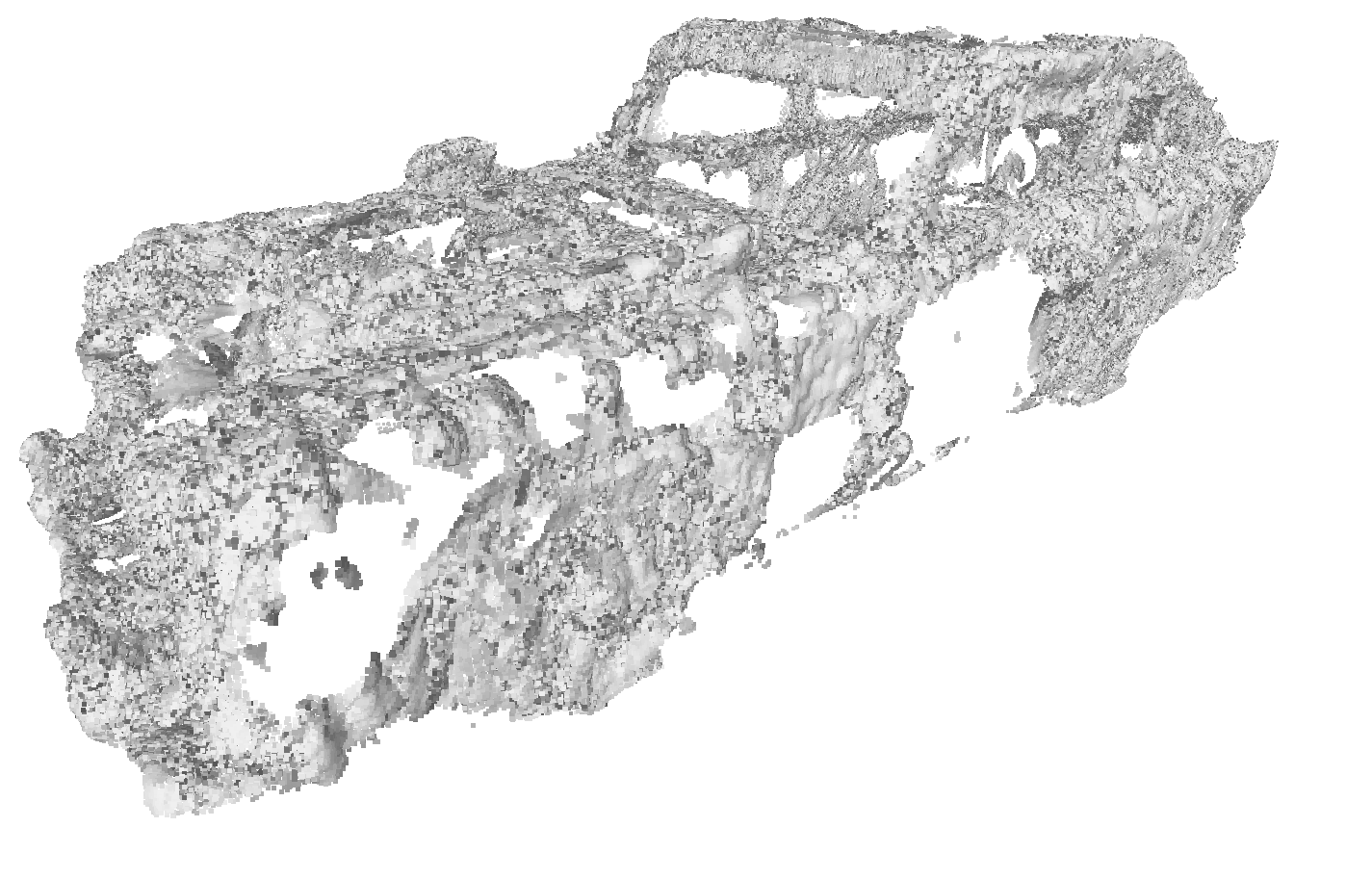}&
        \includegraphics[width=\linewidth,height=2cm,keepaspectratio]{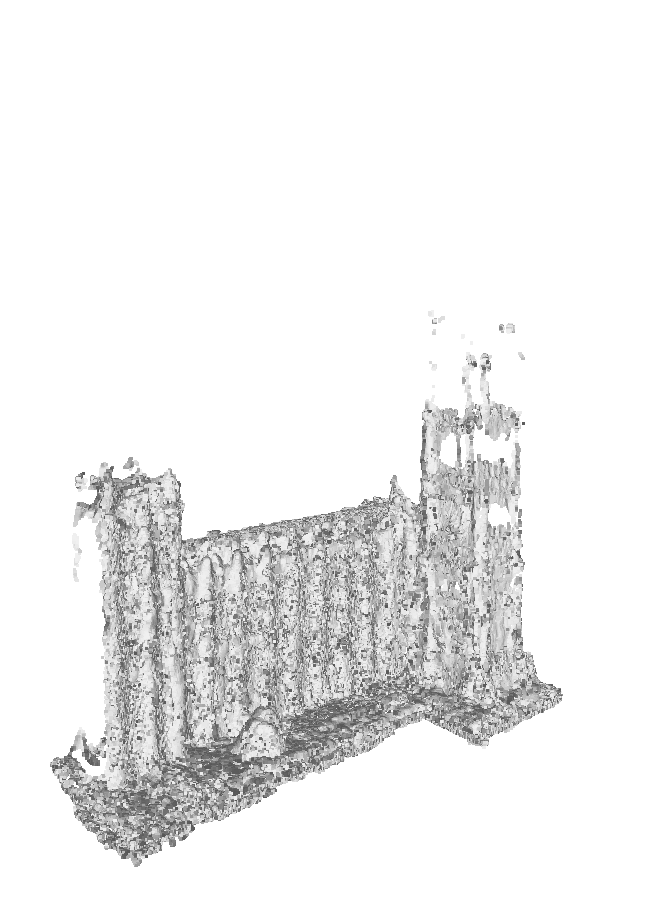}&
        \includegraphics[width=\linewidth]{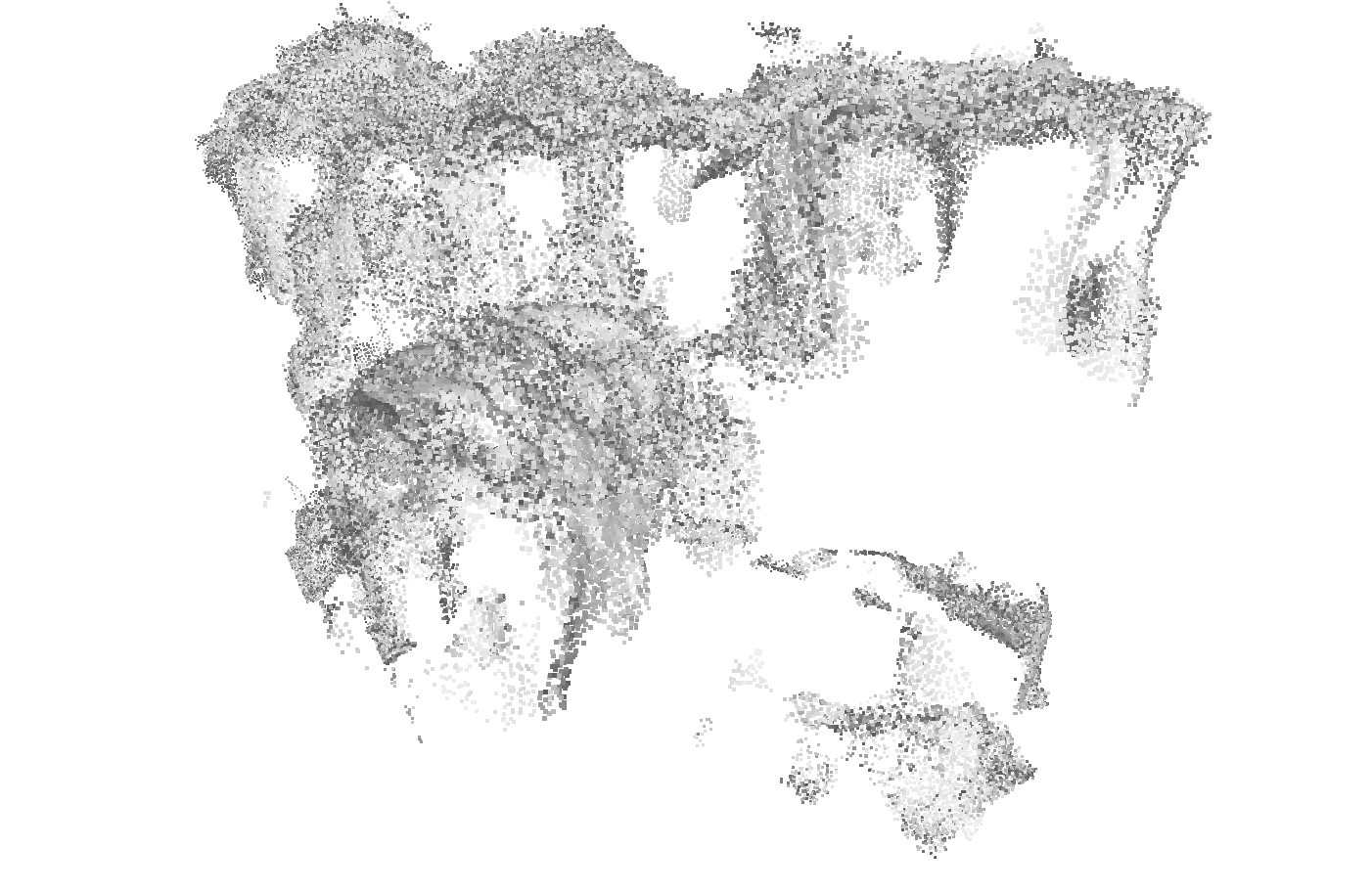}
        \\
        NeRF &
        \includegraphics[width=\linewidth]{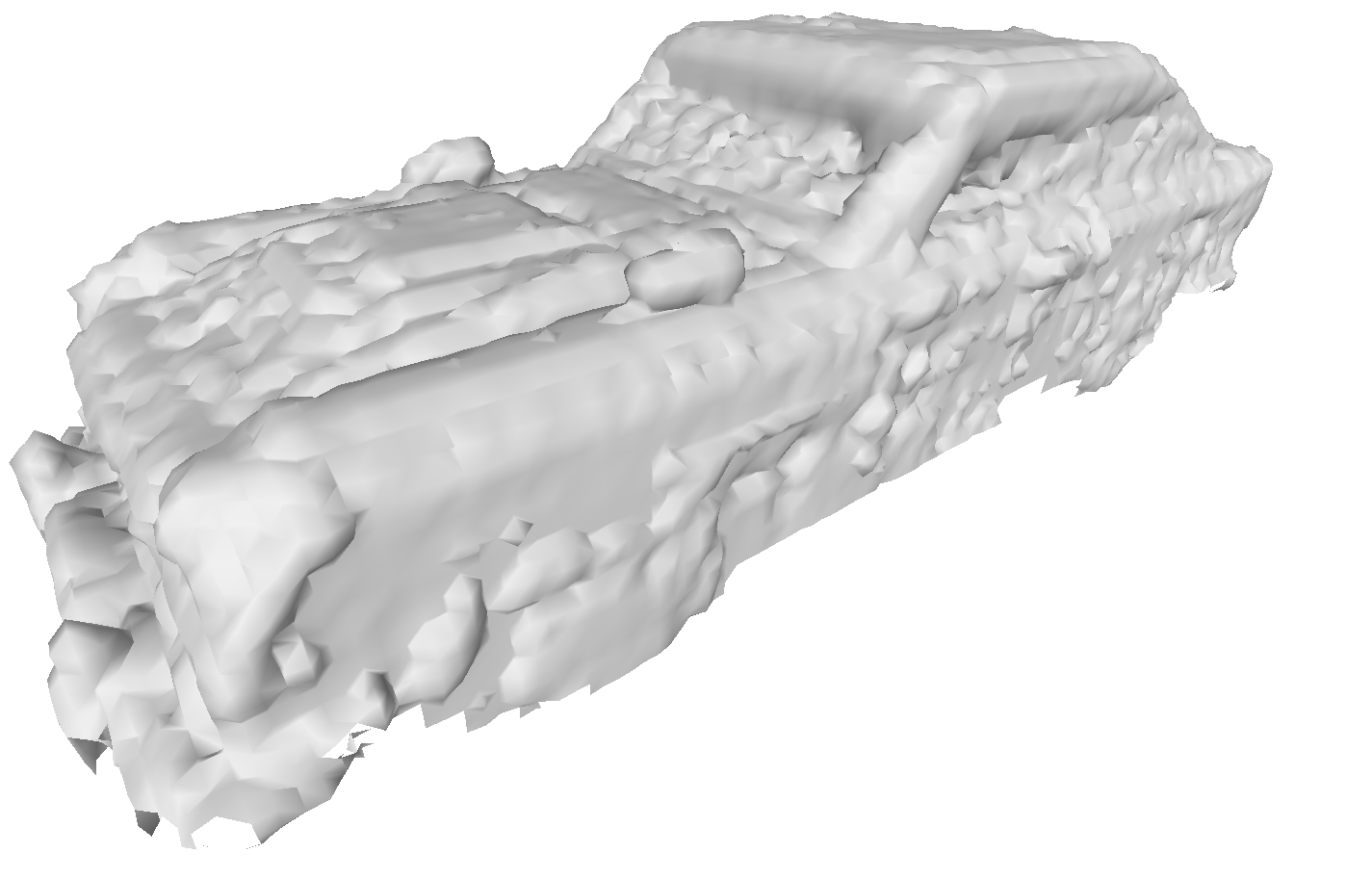}&
        \includegraphics[width=\linewidth,height=2cm,keepaspectratio]{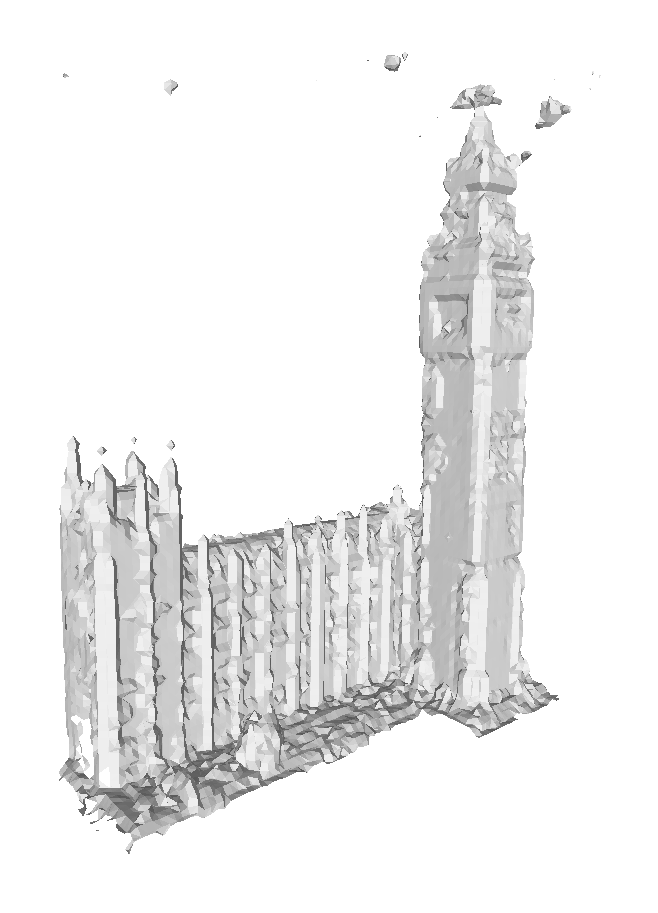}&
        \includegraphics[width=\linewidth]{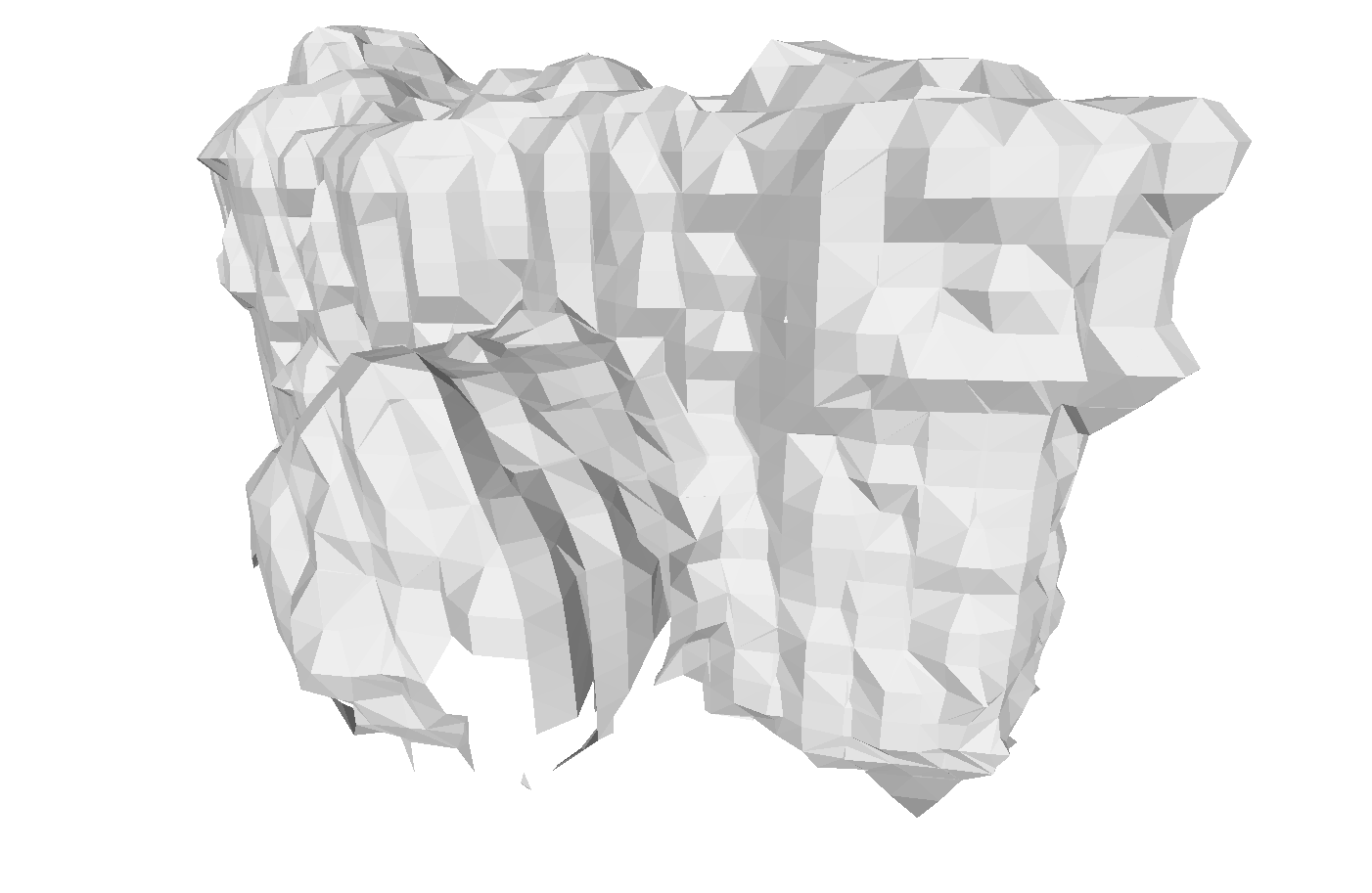}
        \\
        NeuS &
        \includegraphics[width=\linewidth]{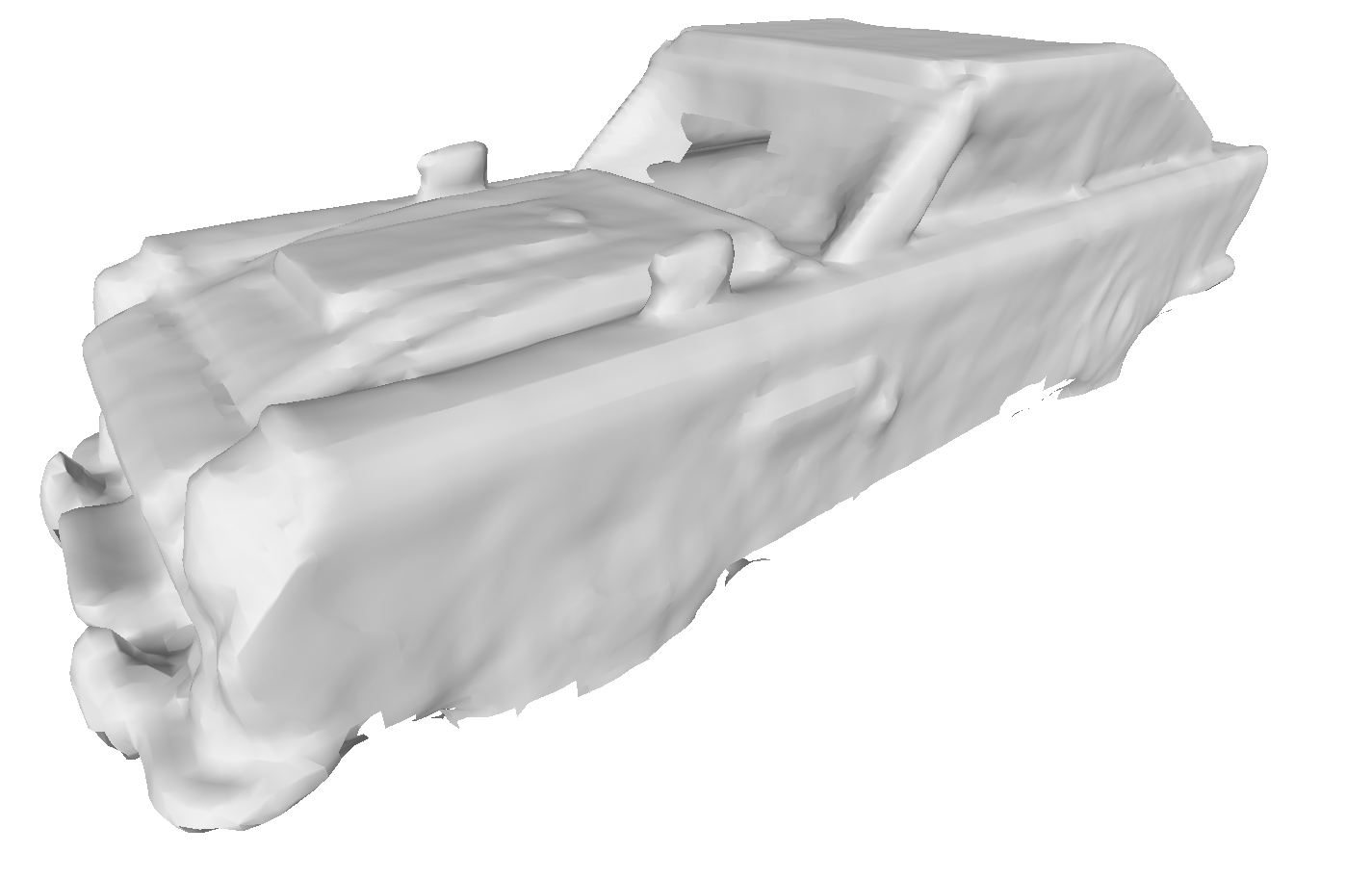}&
        \includegraphics[width=\linewidth,height=2cm,keepaspectratio]{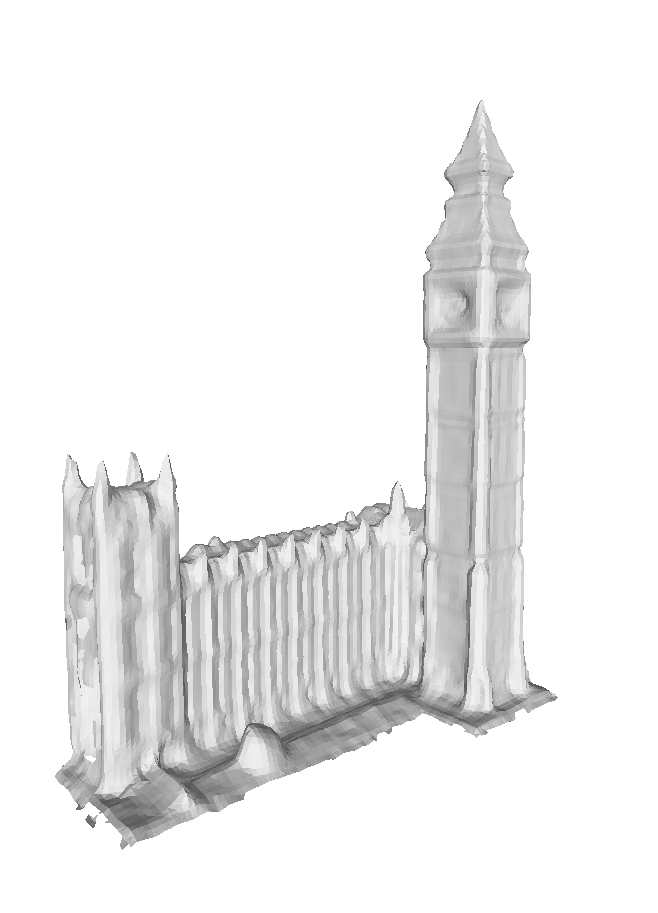}&
        \includegraphics[width=\linewidth]{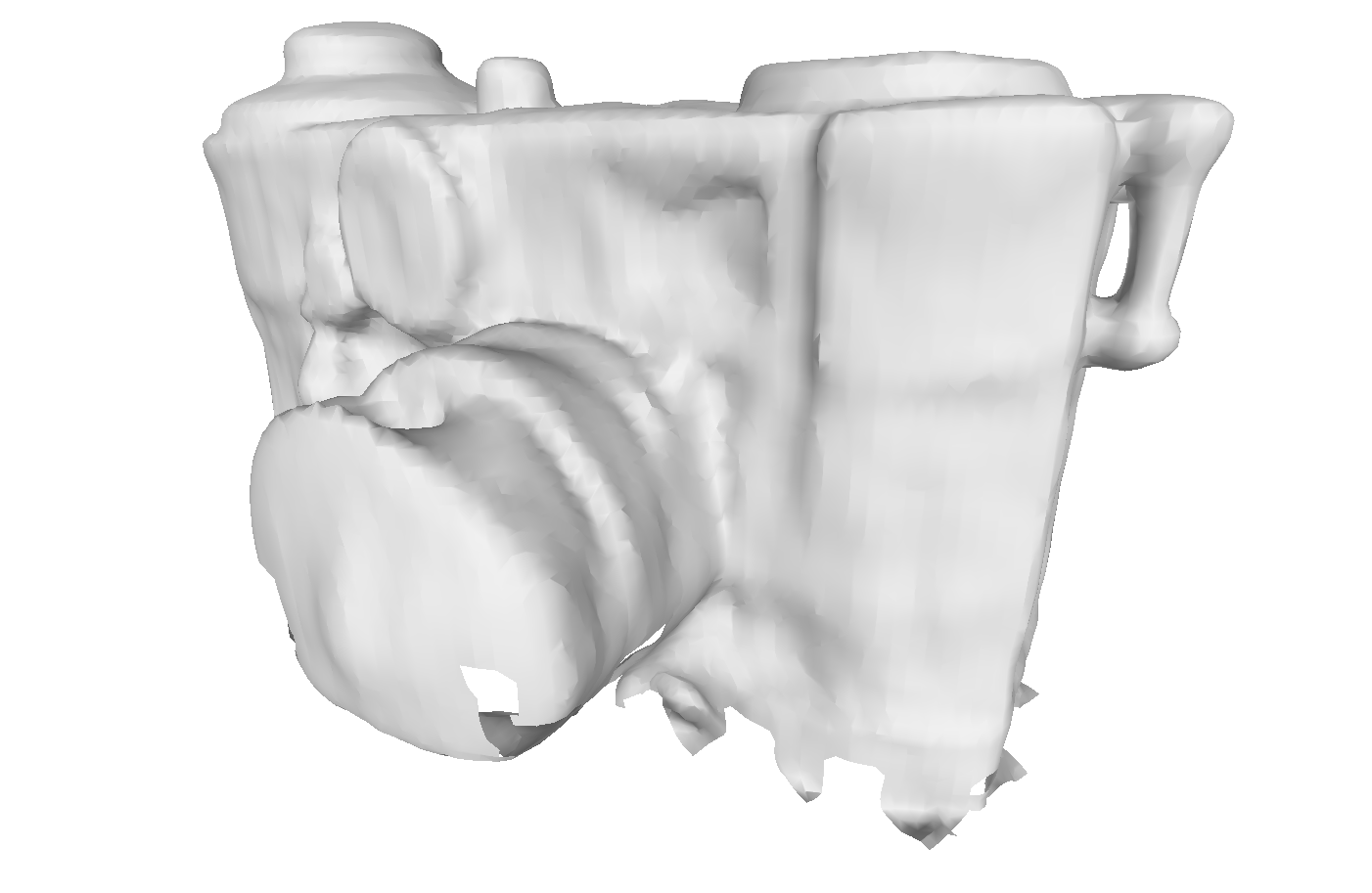}
        \\
        GT &
        \includegraphics[width=\linewidth]{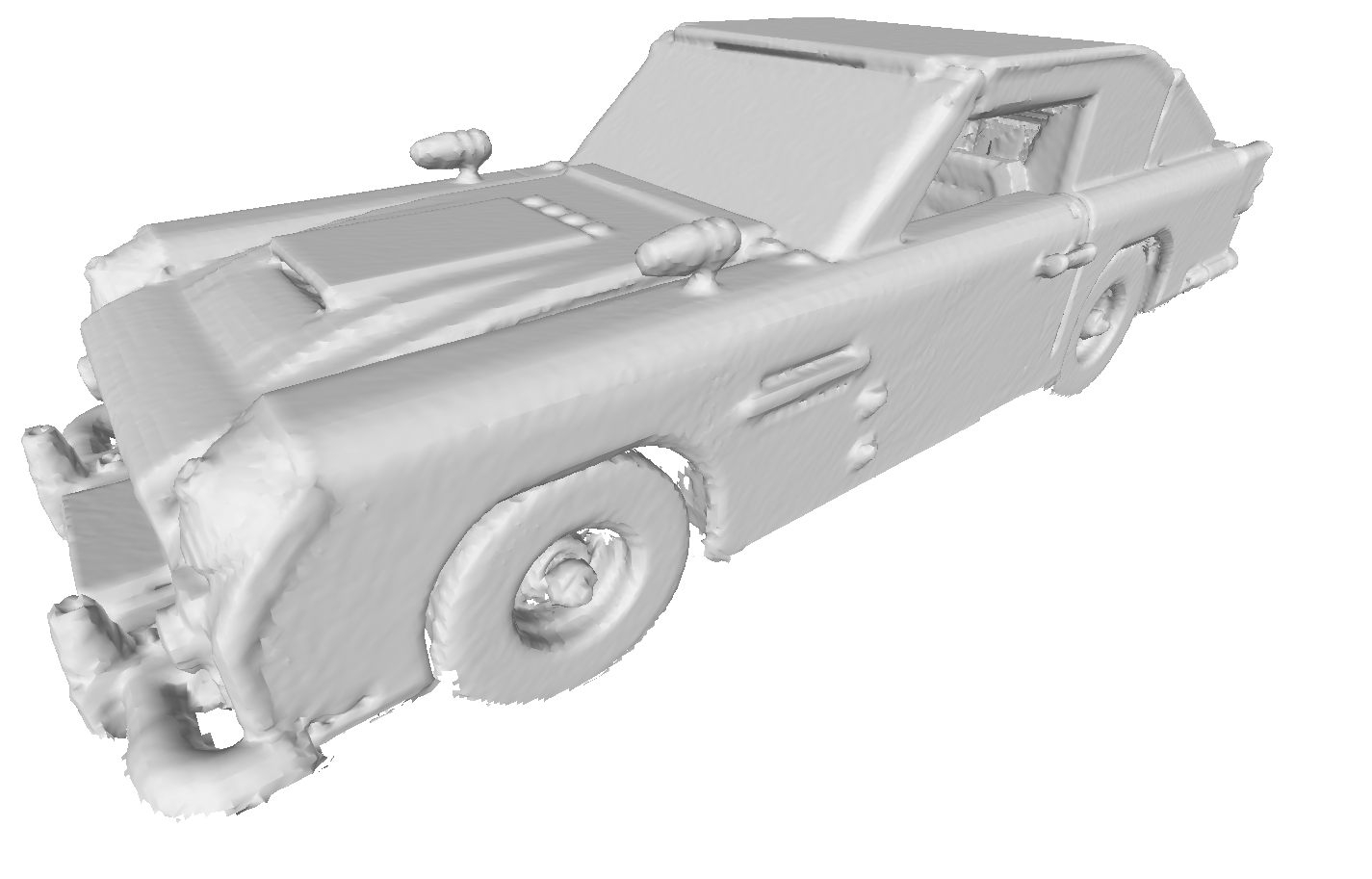}&
        \includegraphics[width=\linewidth,height=2cm,keepaspectratio]{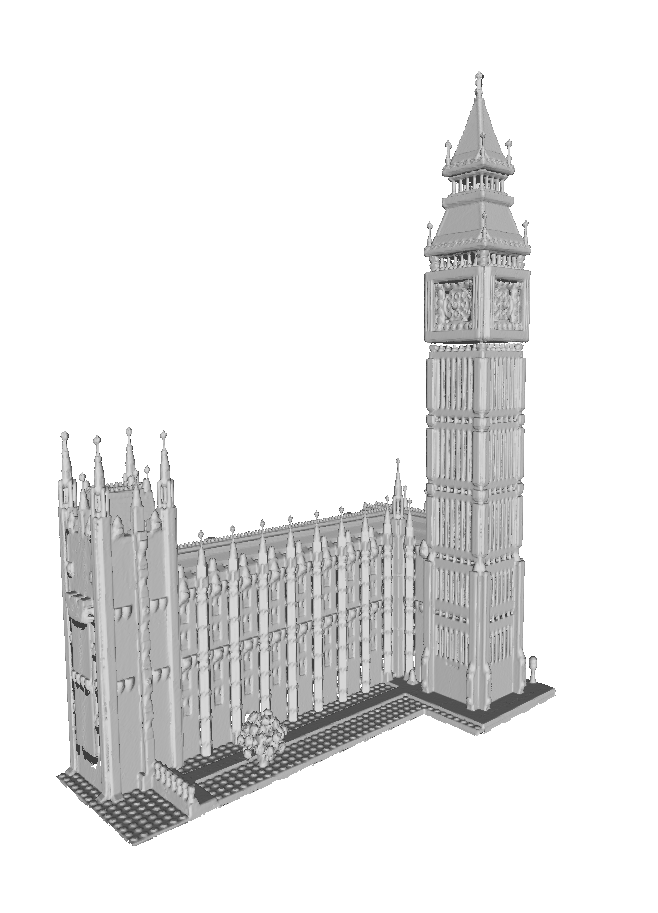}&
        \includegraphics[width=\linewidth]{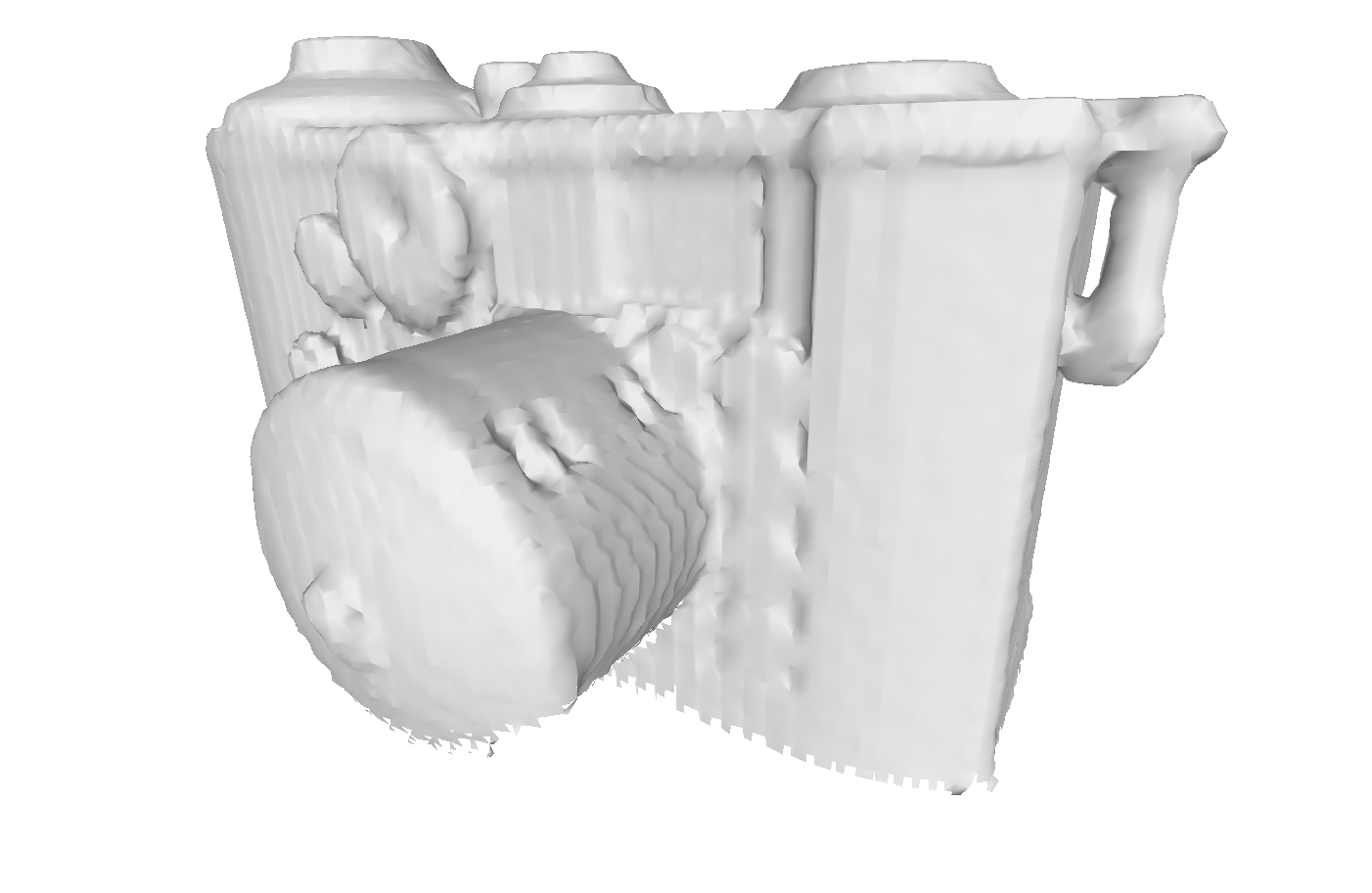}
        \\        
    \end{tabularx}
    
    \caption{Qualitative results of multi-view reconstruction.
\label{fig:mvs}}
\end{figure}

\boldstart{Results} ~\cref{tab:mvs_exp} details the quantitative results, and we visualise some reconstructions in~\cref{fig:mvs}.
NeuS achieves state-of-the-art performance, outperforming traditional MVS, learning-based MVS and depth map fusion approaches. 
However, note that NeuS takes more than $10$ hours to optimise each 3D object.
NeRF, the other representative of the neural field based approach, performs worse than NeuS because it 
represents a scene using volume density rather than SDF as used in NeuS. 
Therefore, NeRF suffers from extracting high-quality surfaces.

MVS solutions, compared to neural-field-based approaches, are penalised heavily in Recall (\ie surface coverage), as they struggle to reconstruct homogeneous regions where texture is insufficient for MVS, as visualised in~\cref{fig:mvs}.
Vis-MVSNet that is finetuned on our random model set performs substantially better than its baseline which is only trained on an external MVS dataset. 
This demonstrates that the knowledge learnt from the random 3D structures can be generalised to real objects.

Although depth maps are beneficial in large-scale 3D reconstruction (\eg reconstructing a living room), the low-resolution depth maps on mobile devices fail to capture geometry details of objects, which causes ethods that take depth maps as input perform the worst in this experiment.
However, it is worth noting that the depth maps are able to fill in homogeneous regions where MVS solutions fail, as shown in~\cref{fig:mvs}.
This suggests that low-resolution depth maps and multi-view high-resolution RGB is complementary for high-fidelity object reconstruction. 


\subsection{Novel View Synthesis}
\boldstart{Setup} 
We compare several neural-field-based approaches for novel view synthesis. Following the evaluation setup in NeRF~\cite{chen2021mvsnerf} for real images, we hold out every $8^{th}$ image in an image sequence for testing.
Besides the methods used in the evaluation of multi-view reconstruction, we also include Direct Voxel Grid Optimisation~\cite{sun2022direct}, a hybrid representation that is able to achieve similar rendering quality as NeRF after training for only a few minutes.

\boldstart{Results}
NeRF and NeuS have comparable performance, as reported in~\cref{tab:nvs_exp}, although NeRF performs significantly worse than NeuS in multi-view surface reconstruction.
DVGO, despite running much faster than NeRF and NeuS ($20$ minutes versus $10$ hours of optimisation), fails to render realistic images in this experiment.

\subsection{Depth Map Enhancement}
\boldstart{Setup} Colour-guided depth map enhancement aims to improve the quality of low-resolution depth maps guided by the corresponding high-resolution RGB images.
We demonstrate that \methodname can be used to train a depth map enhancement network for object scanning in this experiment.
We use the deep neural network architecture introduced in MSPF~\cite{xian2020multi}.
The baseline, denoted as MSPF (baseline) in~\cref{tab:depth_exp}, is trained on the ARKitScene dataset for general scenes.
MSPF (finetuned) is the network that we finetune on the random model set of the proposed dataset.
We render depth maps from the aligned 3D models and use them as ground-truth in training.
Both networks are evaluated on images of the real-world model set.
To provide a point of reference for performance improvement, we also report the results of nearest-neighbour interpolation, which is the most na\"ive approach for depth map upsampling. 
We use standard depth estimation metrics~\cite{eigen2014depth,godard2019digging}.

\boldstart{Results} 
We report the quantitative results in~\cref{tab:depth_exp}.
After finetuning on the random model set, the model improves substantially in all evaluation metrics over the baseline. 
Notably, the improvements from finetuning alone are often larger than the gap between MSPF (baseline) and NN-interpolation.
This demonstrates the random model set is effective for object-centric depth upsampling.

\renewcommand\tabularxcolumn[1]{m{#1}}


\section{Conclusion}

In this paper, we present the \methodname dataset, consisting of a diverse collection of 3D LEGO models and the precise 3D ground-truth annotations associated with RGBD image sequences captured on a mobile device. 
Using this dataset, we establish benchmarks for multi-view reconstruction, novel view synthesis, and depth map enhancement. 
We anticipate that our dataset will provide a valuable resource for researchers investigating multi-view reconstruction using RGBD images from mobile devices, and we believe that our benchmarks will serve as a critical evaluation tool for assessing progress in this field.
However, we acknowledge that the uniform surface property of LEGO bricks can be a limitation of our dataset, which can be alleviated by applying various types of paints on the models in future work.

\boldstart{Acknowledgement}
The authors gratefully acknowledge the financial support provided by Apple. 
This work is also supported by the UKRI grant: Turing AI Fellowship EP/W002981/1 and EPSRC/MURI grant: EP/N019474/1. 
We would also like to thank the Royal Academy of Engineering and FiveAI. 
Furthermore, we extend our gratitude to Shuman Liu for building the LEGO models, and to Wilbur Law for contributing his LEGO models for image capture.

\clearpage

{\small
\bibliographystyle{ieee_fullname}
\bibliography{reference}
}

\end{document}